%% file: aaai-2024.tex
\documentclass[letterpaper]{article} 
\usepackage{aaai24}  
\usepackage{times}  
\usepackage{helvet}  
\usepackage{courier}  
\usepackage[hyphens]{url}  
\usepackage{graphicx} 
\usepackage{natbib}  
\usepackage{caption} 
\usepackage{algorithm}
\usepackage{algorithmic}

\def\eg{\emph{e.g}.} 
\def\ie{\emph{i.e}.} 
\def\etal{\emph{et al}}
\usepackage{newfloat}
\usepackage{listings}
\DeclareCaptionStyle{ruled}{labelfont=normalfont,labelsep=colon,strut=off} 
\floatstyle{ruled}
\newfloat{listing}{tb}{lst}{}
\floatname{listing}{Listing}
\makeatletter
\newcommand{\thickhline}{%
    \noalign {\ifnum 0=`}\fi \hrule height 1pt
    \futurelet \reserved@a \@xhline
}
\makeatother

\usepackage[utf8]{inputenc} 
\usepackage{url}            
\usepackage{booktabs}       
\usepackage{amsfonts}       
\usepackage{nicefrac}       
\usepackage{microtype}      
\usepackage{xcolor}         

\usepackage{amsthm}
\usepackage{amsmath}
\usepackage{amssymb}
\usepackage{graphicx}
\usepackage{multirow}
\usepackage{dsfont}
\usepackage{subcaption}
\usepackage{capt-of}
\usepackage{lipsum,booktabs}

\newcommand{\mytilde}{\raise.17ex\hbox{$\scriptstyle\mathtt{\sim}$}}
\newtheorem{theorem}{Theorem}

\title{HISR: Hybrid Implicit Surface Representation for Photorealistic 3D Human Reconstruction}

\author{
    Angtian Wang\textsuperscript{\rm 1}\footnote{The work is done during Angtian Wang's internship at Meta. The corresponding author is Yuanlu Xu.}, Yuanlu Xu\textsuperscript{\rm 2}, Nikolaos Sarafianos\textsuperscript{\rm 2}, Robert Maier\textsuperscript{\rm 2}, Edmond Boyer\textsuperscript{\rm 2}, Alan Yuille\textsuperscript{\rm 1}, Tony Tung\textsuperscript{\rm 2}\\
}
\affiliations{
    \textsuperscript{\rm 1} Johns Hopkins University, \;\;
    \textsuperscript{\rm 2} Meta Reality Labs Research \\

}

\usepackage{bibentry}

\begin{document}

\maketitle

\begin{abstract}


Neural reconstruction and rendering strategies have demonstrated state-of-the-art performances due, in part, to their ability to preserve high level shape details. Existing approaches, however, either represent objects as implicit surface functions or neural volumes and still struggle to recover shapes with heterogeneous materials, in particular human skin, hair or clothes. To this aim, we present a new hybrid implicit surface representation to model human shapes. This representation is composed of two surface layers that represent opaque and translucent regions on the clothed human body. We segment different regions automatically using visual cues and learn to reconstruct two signed distance functions (SDFs). We perform surface-based rendering on opaque regions (\eg\,body, face, clothes) to preserve high-fidelity surface normals and volume rendering on translucent regions (\eg\,hair). Experiments demonstrate that our approach obtains state-of-the-art results on 3D human reconstructions, and also shows competitive performances on other objects. 




\end{abstract}

\section{Introduction}
\label{sec:intro}

Realistic and accurate reconstruction of geometry and appearance of digital humans has received significant attention over the past few years with applications ranging from creating vivid characters to virtual assistants in customer service, and social telepresence~\cite{lombardi2018deep}.

 Exist techniques recover the geometry and appearance of humans from images (\eg\,using multi-view stereo, shape-from-X, etc.). 
 Following the introduction of NeRF~\cite{nerf20}, which employs a neural network to capture color and opacity in a 3D volume, methods using neural radiance fields have gained significant popularity over the past few years.
 The NeRF network learns to estimate the radiance optimally from any viewpoint, resulting in images with photorealism. However, NeRF and the related works struggle to accurately capture fine-level surface details and may generate erroneous discrete floating volumes. 
Signed distance fields (SDF)~\cite{DeepSDF19},  that model the closest distances to surfaces, have been proposed as an alternative to opacity in implicit shape representations. Their advanced ability to help recover better geometries has been demonstrated in \eg \cite{IDR20, VolSDF21, UNISURF21}.


Our experiments show that concurrent SDF-based representations lack the ability to model fine structures and high-frequency regions of humans (\eg, hair and complex cloth patterns). While NeRF-like representations are too sensitive to noise and misalignment, and tend to generate large amount of floating volumes on the final reconstruction. 
Such differences potentially stem from, first, the nature of implicit representation with distance fields which provides a smooth transition from positive to negative values across the surface boundary. SDFs are thus less likely to produce floating volumes, which requires change of the surface direction in high frequency. Second, the Eikonal loss $\|\nabla \Phi(\mathbf{x})\|-1$, which further enforces the smoothness of object boundaries, and reduces therefore the sharpness and floating volumes.



\begin{figure*}
\centering
\includegraphics[width=0.98\textwidth]{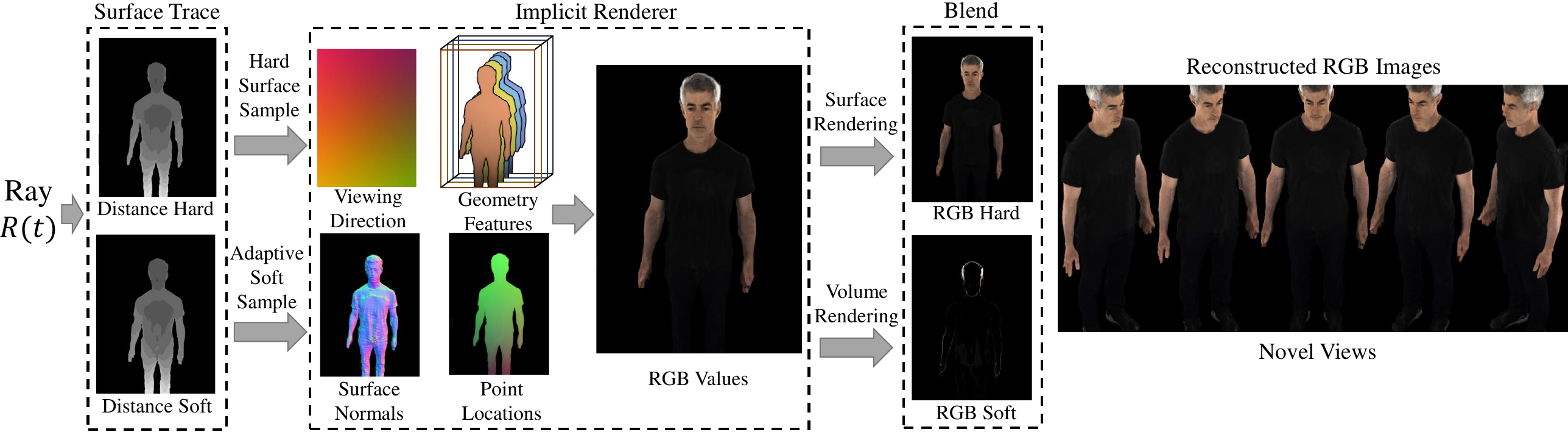}
\caption{\emph{Overview of HISR}, which takes viewing rays as input and simultaneously conducts surface and volume rendering. The process is automatically controlled by two SDFs. Once each rendering is done, we blend the colors for the final output. Due to privacy reason, we mask out part of the face. }
\label{fig:pipeline}
\end{figure*}

Previous works~\cite{UNISURF21,VolSDF21,NeuS21} attempt to combine SDF and NeRF by replacing density fields with SDFs and employing SDF-to-density functions, while performing volume rendering by sampling the entire space. These approaches offer advantages such as enforcing surface smoothness through geometry regularization and enabling training without a segmentation mask. However, we observe that such models lose the crucial capability of NeRF models to capture intricate geometries, like hair strands, and struggle to converge under challenging scenarios. In this paper, we propose a complementary approach that retains the strengths of both SDF and NeRF models by carefully controlling the model's behavior according to different body parts, allowing for improved performances and the ability to capture detailed geometries.


In this paper, we propose a new neural rendering framework for multi-view reconstruction of real humans. We use the signed distance field (SDF) to model shape surfaces and introduce a novel volume rendering scheme to learn a two-layer implicit surface-based representation. Specifically, by introducing a density distribution induced by the SDF, we can perform volume rendering to learn an implicit SDF representation and thus obtain both an accurate surface representation, benefiting from the neural SDF model, and a robust network training in the presence of abrupt depth changes, as enabled by the volume rendering.
We performed in-depth quantitative evaluations of our Hybrid Implicit Surface Representation (HISR) on stage-captured real people and photo-realistic virtual humans, as well as objects from the DTU dataset \cite{aanaes2016large}. We demonstrate that our method is capable of reconstructing photo-realistic 3D clothed humans and clearly outperforms several state-of-the-art approaches quantitatively and qualitatively.
Our contributions include: 
\begin{itemize}
    \item A study of state-of-the-art 3D reconstruction approaches for 3D humans, which shows \emph{existing approaches cannot obtain both smooth surface reconstruction and high-fidelity geometry details simultaneously}.
    \item \emph{A new hybrid representation}, which is surface-based while enabling volume rendering for fine-grained geometry.
    \item A \emph{computed expectation of SDF values within a conical frustum} when computing the volume densities by considering the viewing rays as cones, which significantly improves the reconstructed geometry.
    \item \emph{A new loss to regularize the specularity} changes upon viewing directions.
\end{itemize}



\section{Related Work}
\label{sec:literature}


\textbf{Neural Implicit Representations} based on ray tracing volume densities \cite{raytracevol84}  formulate the volume rendering process as the solution for the scatter equation under the low albedo assumption. Different from early approaches on volume rendering, which represent objects using explicit primitives \cite{westover1990footprint, zwicker2001ewa, wang2022voge}, NeRF~\cite{nerf20} represents objects via implicit functions of volume densities, which has shown high-quality results in the novel view synthesis task. 
Follow-up works \cite{mipnerf21, barron2022mipnerf360} further improve the novel view synthesis ability of NeRF with more fine-grained details. 
Recent works explore broader applications of NeRF \cite{chen2022hallucinated, xu2022point, wang2022rodin, gao2023surfelnerf, cai2023structure, wang2023benchmarking}.
However, the reconstructed geometry of NeRF results in artifacts since the geometry representation lacks surface constraints.
Orthogonal to the above studies, works focus on applying implicit representations to human sequences~\cite{InsNGP22,neus222,humanrf23}. 

\textbf{Neural 3D Human Reconstruction} has shown great potential in many digital human and AR/VR applications~\cite{mescheder2019occupancy,park2019deepsdf,chen2019learning,huang2020arch}. 
Different from regular objects \cite{wang2021nemo, wang2023neural}, the geometries of human are articulate with large variant appearance on different regions. 
One of the first approaches to adopt the implicit function representation for 3D human reconstruction from a single image is PIFu~\cite{PIFuICCV19, saito2020pifuhd}. PIFu leverages pixel-aligned image features rather than global features. Local details present in the input image are preserved as the occupancy of any 3D point is predicted. 
Alldieck \etal~\cite{alldieck2022photorealistic} improved upon PIFu by introducing a network that estimates 3D geometry, surface albedo and shading from a single image in a joint manner.
In contrast, inspired by the literature of image-based super-resolution, SuRS~\cite{SuRSECCV2022} demonstrates that fine-scale detail can be recovered even from low-resolution input images using a multi-resolution learning framework.
While other approaches using NeRF-type approach for reconstruct 3D human in motion \cite{pumarola2021d,gafni2021dynamic,NeuralBody21,Narf21, jiang2022alignerf, weng2022humannerf}.

\begin{figure*}
\centering
\begin{subfigure}{.59\textwidth}
\includegraphics[width=\textwidth]{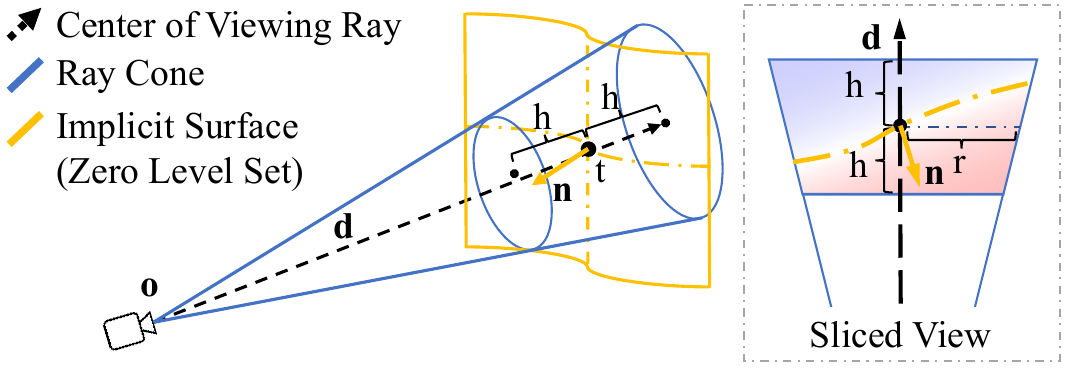}
\caption{}
\label{fig:ray_cone}
\end{subfigure}
\vline
\begin{subfigure}{.39\textwidth}
\includegraphics[width=\textwidth]{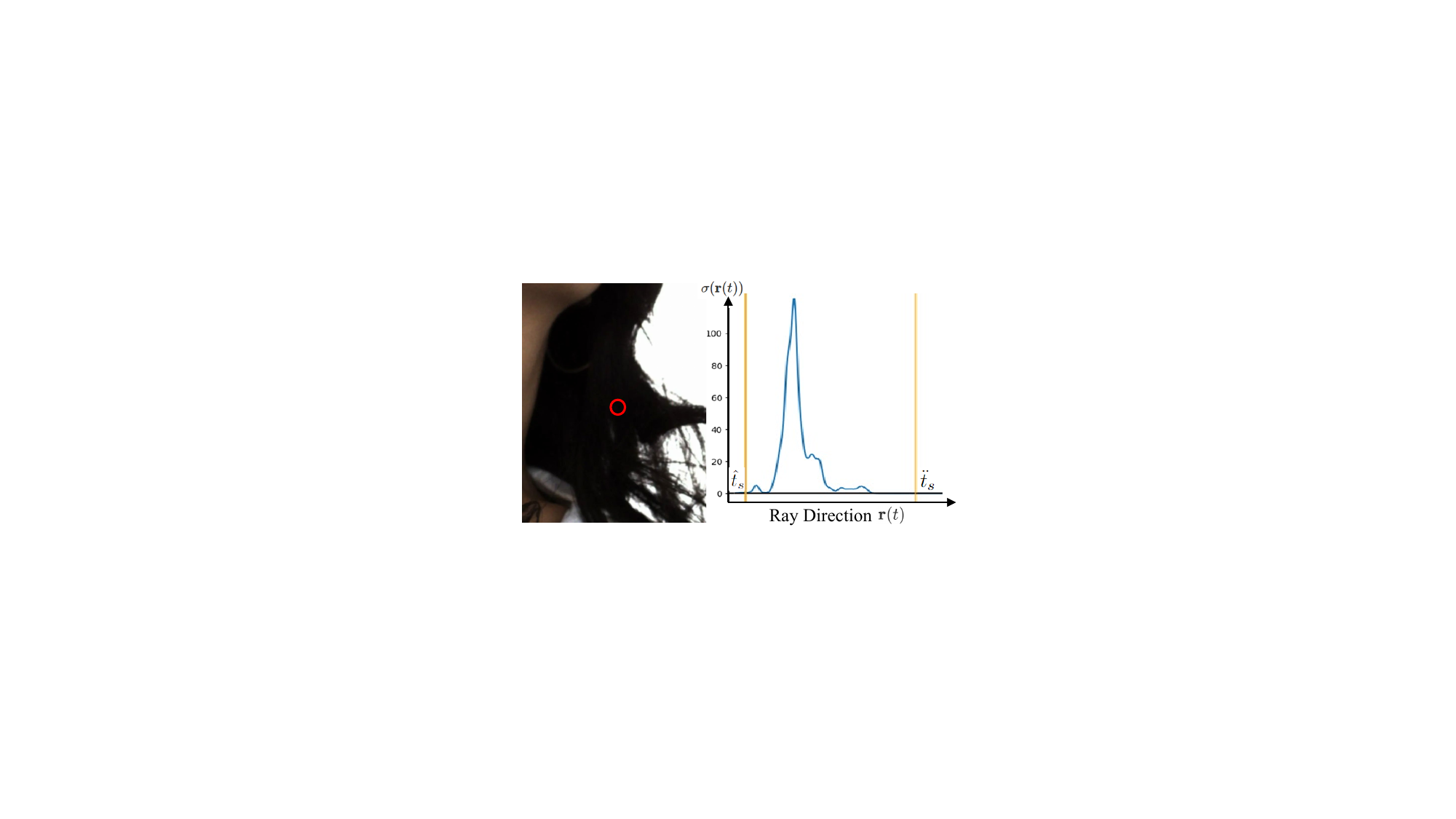}
\caption{}
\label{fig:density_hair}
\end{subfigure}
\caption{(a) \textit{Illustration of cone cast for volume rendering.} We assume a uniform SDF inside the conical frustum and compute a close-form solution for the expectation of the SDF values. (b) \textit{Examples of volume densities computed from SDF values on hair regions.} The ray is sampled at the center of the red circle. We conduct volume rendering from $t=\hat{t}_s$ to $t = \ddot{t}_s$. }
\label{fig:ray_cone_and_density}
\end{figure*}

\textbf{Geometry Representation of Signed Distance Function} allows for modeling objects by an implicit surface~\cite{DeepSDF19}. IDR~\cite{IDR20} proposed to use a single implicit hard surface as object representation and achieved high-quality geometry reconstructions. Follow-up works~\cite{VolSDF21,NeuS21,yariv2023bakedsdf} learn and render SDF functions in a volume rendering manner, which demonstrate comparable better reconstruction for the smooth surface without mask supervision. 
However, compared to general objects for which getting very accurate segmentation or matting masks might be challenging, getting high-quality human masks is a well-researched topic. 
Powerful pre-trained segmentation models for clothed humans allow us to obtain the mask at almost no cost.
On the other hand, such approaches lose the ability to model geometries with fine-level details, while in this work, we follow the SDF-based approach for geometry reconstruction but with the capacity for modeling detailed geometries.


\input{Method}

\input{Experiment}

\section{Conclusion}
\label{sec:conclusion}

In this paper, we introduced HISR, a novel hybrid implicit surface representation for photo-realistic 3D human reconstruction, employing a unique volume rendering scheme to maintain surface detail and realism. Our method computes the expected Signed Distance Function (SDF) values within a conical frustum, enhancing the quality of reconstructed geometry.
Evaluated across various human and object reconstruction datasets, our approach surpasses baseline methods in both reconstructed geometry fidelity and novel view synthesis. This is due to our dual surface layer representation for opaque and translucent regions, allowing for nuanced rendering of complex human features like skin, hair, and clothing.
In conclusion, our hybrid representation significantly advances neural reconstruction and rendering, particularly in handling heterogeneous geometries. Our approach achieves state-of-the-art results in 3D human reconstructions and shows promise in other objects.

\noindent {\small{\textbf{Acknowledgements.}} We would like to thank Ziyan Wang for the help of running comparison experiments and tuning WIDC camera calibration.}



\bibliography{aaai24}
\newpage

\input{AppendixS}

\end{document}

%% file: Method.tex
\section{Method}
\label{sec:method}

Our approach builds on two key components: i) signed distance functions (SDF)~\cite{DeepSDF19} for representing the instance's geometry, and ii) an implicit neural renderer capable of modeling textures and illumination. To elucidate our method, we first introduce the two primary rendering mechanisms used for implicit representations: surface-based and volume-based. 
And to better integrate the surface and volume rendering approaches, we introduce a novel technique called integrated SDF for volume densities. This approach enables the seamless fusion and synchronization of updated SDF and volume densities within our approach, enhancing the overall quality and accuracy of the reconstructed human model.

\subsection{Hybrid Implicit Surface Representation}

 
\textbf{Surface Rendering.} The 3D object is represented as a set of points on surface, which is described as the zero level set of the SDF $\Phi$ represented by a neural network (MLP),
\begin{equation}\small
\mathcal{S}=\left\{\mathbf{x} \in \mathbb{R}^3 \mid \Phi(\mathbf{x})=0\right\}.
\end{equation} 
During the rendering process, we compute a ray $\mathbf{r}(t) = \mathbf{o} + t \cdot \mathbf{d}$ for a given image pixel, where $\mathbf{o}$ represents the camera location, $\mathbf{d}$ is the viewing direction and $t$ is the depth along the viewing ray. To determine the intersection of the ray $\mathbf{r}(t)$ with the surface $\mathcal{S}$, we use surface tracing to search along the ray for the first zero point, denoted by $\hat{t}$, where $\Phi(\mathbf{r}(\hat{t})) = 0$. The final color observation is computed at the intersection point as:
\begin{equation}\small
\textbf{C} = M(\hat{\mathbf{x}}, \hat{\mathbf{n}}, \mathbf{d}, f),
\label{equ:implicit_renderer}
\end{equation}
where $\hat{\mathbf{x}} = \mathbf{r}(\hat{t})$ is the point location, $\hat{\mathbf{n}}$ is the surface normal at $\hat{\mathbf{x}}$, $M$ denotes the implicit neural renderer implemented with a MLP, and $f$ is a feature vector describing the geometry that is obtained from the SDF network.

\noindent \textbf{Volume Rendering.} The 3D object is represented as semi-transparent volumes with volume density $\sigma(\mathbf{x})$ at each location in the scene $\mathbf{x} \in \mathbb{R}^3$. During the rendering process, for each viewing ray $\mathbf{r}(t) = \mathbf{o}+t\cdot \mathbf{d}$, a set of points $\mathbf{r}(t_k)$ is sampled and stored, along with the computed color $\mathbf{c}_k$ and density $\sigma_k$. The final color observation is obtained by discretizing the sum, approximating the integral within the volume rendering function~\cite{raytracevol84,nerf20}:
\begin{equation}
\begin{aligned}\small
\mathbf{C}=\sum_k T_k\left(1-\exp \left(-\sigma_k\left(t_{k+1}-t_k\right)\right)\right) \mathbf{c}_k, \\
\text { with }  T_k=\exp \left(-\sum_{k^{\prime}<k} \sigma_{k^{\prime}}\left(t_{k^{\prime}+1}-t_{k^{\prime}}\right)\right),
\label{equ:volume_rendering}
\end{aligned}
\end{equation}
where $T_k$ is the transmittance function which encodes the visibility at each sampled point. In order to conduct volume rendering using an SDF as the geometry representation, a signed distance to volume density function $\Phi$ is introduced:
\begin{equation}
    \sigma(\mathbf{x}) = \Psi(-\Phi(\mathbf{x})),
\end{equation}
where $\Psi$ is the derivative of the Cumulative Distribution Function (CDF) of Laplace~\cite{VolSDF21} or Gaussian~\cite{NeuS21} distribution.

\noindent \textbf{Hybrid Implicit Surface Representation.} In our proposed approach HISR, the geometry is represented as a set of surfaces with volumes filled in between specific surfaces. We assume that the space inside the hard surface $\mathcal{P}_h$ is filled with opaque materials, while the outside volumes are translucent with volume densities $\sigma(\mathbf{x})$. Specifically, the boundary of the translucent region is determined by the hard surface, as the inner boundary, and another surface as the outer boundary namely the soft surface $\mathcal{P}_s$. Whereas the space outside the soft surface is vacant.
The inner and outer surfaces are represented by two SDFs, namely hard SDF and soft SDF. The hard $s_h$ and soft SDF values $s_s$ at each location $\mathbf{x}$ in the space $\mathbf{x} \in \mathbb{R}^3$ are:
\begin{equation}
    s_h = \Phi_h (\mathbf{x}),\;\; s_s = \Phi_s (\mathbf{x}),
\end{equation}
where $\Phi_h$ and $\Phi_s$ are two outputs generated by a shared MLP. In HISR, similar to other SDF-based volume rendering approaches, the volume density is represented by the SDF-to-density function $\sigma(\mathbf{x}) = \Psi(\Phi(\mathbf{x}))$. 

\noindent \textbf{Hybrid Rendering.} Figure~\ref{fig:pipeline} illustrates the rendering process of HISR. Given a viewing ray $\mathbf{r}(t) = \mathbf{o}+t\cdot \mathbf{d}$, we perform surface tracing to find the intersection point on the ray with $\mathcal{P}_h$ and $\mathcal{P}_s$. This process involves searching for the zero level along $\mathbf{r}(t)$ to obtain $\hat{t}_s$ and $\hat{t}_h$ such that $\Phi_s(\mathbf{r}(\hat{t}_s)) = 0$ and $\Phi_h(\mathbf{r}(\hat{t}_h)) = 0$. Consequently, we can deduce that the ray passes through a vacancy from the camera location at $t=0$ to $t=\hat{t}_s$, which contributes nothing to the final viewing color. From $t=\hat{t}_s$ to $t=\hat{t}_h$, the ray traverses translucent regions, and the final viewing color $\mathbf{C}_s$ is computed using Eq.~\ref{equ:volume_rendering} (the color network is shared with the outer surface rendering). At $t=\hat{t}_h$, the ray intersects with $\mathcal{P}_h$, and the color $\mathbf{C}_h$ can be computed using Eq.~\ref{equ:implicit_renderer}. Beyond $t=\hat{t}_h$, the ray becomes entirely invisible as the transmittance $T(\mathbf{r}(t))$ reduces to $0$. The final viewing color is calculated as $\mathbf{C} = \mathbf{C}_s + T(\hat{t}_h) \cdot \mathbf{C}_h$.

\begin{figure}
   \includegraphics[width=0.47\textwidth]{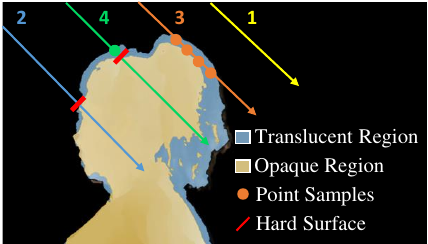}
    \caption{HISR conducts volume rendering in translucent regions and surface rendering on hard surface. The numbers refer to the rendering cases presented in the table below.}
    \label{fig:ray_sample}
\end{figure}

Indeed, the viewing ray $\mathbf{r}(t)$ may only intersect one of the surfaces or even none of them. In the case where $\mathbf{r}(t)$ intersects $\mathcal{P}_s$ but not $\mathcal{P}_h$, we search for a secondary intersection of $\mathbf{r}(t)$ with $\mathcal{P}_s$ such that $\Phi_s(\mathbf{r}(\ddot{t}_s)) = 0$, where $\ddot{t}_s > \hat{t}_s$. This allows the ray to pass inside the volume from $t=\hat{t}_s$ to $t=\ddot{t}_s$ before encountering a vacancy. Figure \ref{fig:ray_sample} illustrates all possible cases, which summarize as follows:

\begin{center}
\small
\begin{tabular}{@{}c|c|c|c|c@{}}
 \hline
  $\mathbf{r}(t) \cap \mathcal{P}_s$ & $\mathbf{r}(t) \cap \mathcal{P}_h$ & start & end & final color \\
 \hline
   $= \emptyset$ & $= \emptyset$ & --- & --- & $\mathbf{C}_{bg}$ \\
  $= \emptyset$ & $\neq \emptyset$ & --- & --- & $\mathbf{C}_{h}$ \\
   $\neq \emptyset$ & $= \emptyset$ & $t=\hat{t}_s$ & $t = \ddot{t}_s$ & $\mathbf{C}_s + T(\ddot{t}_h) \cdot \mathbf{C}_{bg}$ \\
  $\neq \emptyset$ & $\neq \emptyset$ & $t=\hat{t}_s$ & $t = \hat{t}_h$ & $\mathbf{C}_s + T(\ddot{t}_h) \cdot \mathbf{C}_{h}$ \\
 \hline
\end{tabular}
\end{center}

Additionally, we observed that sampling the same number of points on rays belonging to cases 3 and 4 is inefficient. This is because rays in case 4 typically have a much larger sampling range, and the overall rendering quality primarily relies on the largest sampling interval. To mitigate this issue and minimize the maximum sampling interval while staying within the constraints of limited GPU memory, we propose an \textit{Adaptive Sampling Strategy}. This strategy allows for different numbers of points to be sampled on rays while ensuring that the sampling interval on each ray remains similar. We implement this strategy using CUDA as differentiable PyTorch functions. For more details refer to Section \ref{sec:ada_sample}.

\subsection{Integrated SDF for Volume Densities}
\label{sec:method:density}

\textbf{Gaussian Mixture for SDF-to-density Function.} The density of semi-transparent volumes is determined using the SDF-to-density functions $\sigma(\mathbf{x}) = \Psi(-\Phi(\mathbf{x}))$. However, we have observed that the $\Psi$ function used in previous works~\cite{VolSDF21,NeuS21}  yields a smooth surface while being successful at capturing intricate geometric details.
To address this issue, we introduce the learnable Gaussian mixture model for SDF-to-density:
\begin{equation}\small
    \sigma(\mathbf{x}) = \sum_{k=0}^K \alpha_k \cdot \exp(-\frac{(\Phi(\mathbf{x}) - \mu_k)^2}{\beta_k^2}),
\label{equ:gmm_density}
\end{equation}
where $\alpha_k$, $\beta_k$ and $\mu_k$ are learnable parameters for k-th Gaussian mixtures. Figure \ref{fig:density_hair} shows how our Gaussian mixture functions allows to model fine-grained details as well as successfully reconstruct hair.

\noindent \textbf{Integrated SDF with Cone Cast.} 
However, unlike previous approaches that solely perform volume rendering, we have identified inconsistencies and visible artifacts on the geometry during hybrid rendering, as can be observed in the bottom-right visualization of Figure~\ref{fig:widc_normal}. These artefacts result from inaccuracies in estimating volume densities near the outer surface. Inspired by mip-NeRF~\cite{mipnerf21}, we propose to consider the ray as a cone rather than a single infinitesimally narrow line within the volume rendering region. To this aim, we investigate the evaluation  of the SDF expected value within a conical frustum (a section of the cone) for each sample.

As Figure~\ref{fig:ray_cone} shows, the apex of that cone lies at $\mathbf{o}$ with the axis along the viewing ray. The radius of the cone at location $\mathbf{o} + t \cdot \mathbf{d}$ is $r$. The set of positions $\mathbf{x}$ within a conical frustum between $[ t - h, t + h ]$ is:
\begin{equation}
\begin{aligned}
\small
\mathrm{F}(\mathbf{x}, \mathbf{o}, \mathbf{d}, r, t, h) = \mathds{1}\bigg\{ & (t - h < \mathbf{d}^{\mathrm{T}}(\mathbf{x} - \mathbf{o}) < t + h) \\
& \wedge \left(\frac{\mathbf{d}^{\mathrm{T}}(\mathbf{x} - \mathbf{o})}{\|\mathbf{x} - \mathbf{o}\|_2} > \frac{t}{\sqrt{t^2 + r^2}}\right) \bigg\},
\end{aligned}
\end{equation}
where $\mathds{1}$ is an indicator function: $F(\mathbf{x}, \mathbf{\cdot})$ iff $\mathbf{x}$ is
within the conical frustum defined by $(\mathbf{o}, \mathbf{d}, r, t, h)$.

\begin{theorem}\label{thm:distance}
 Inside a uniform signed distance field, where gradient directions on each location are parallel, given a known SDF value $s_c$ at location $\mathbf{x}_c$ and gradient direction $\mathbf{n}_c$, the SDF value $s$ at any location $\mathbf{x}$ is computed as 
\begin{equation}\small
    s = \Phi(\mathbf{x})=s_c+\mathbf{n}_c \cdot (\mathbf{x}_c - \mathbf{x}).
\end{equation}
\end{theorem}
The proof of this theorem is provided in Section \ref{sec:app:proof}. By assumption of the uniform SDF within each conical frustum, we compute the expected SDF values following Theorem \ref{thm:distance}:
\begin{equation}\small
    \mathrm{E}[s] = \frac{\int \Phi(\mathbf{x}) \cdot \mathrm{F}\left(\mathbf{x}, \mathbf{o}, \mathbf{d}, r, t, h\right) d\mathbf{x}}{\int \mathrm{F}\left(\mathbf{x}, \mathbf{o}, \mathbf{d}, r, t, h\right) d\mathbf{x}}.
\end{equation}
Here we compute both $\mathrm{E}[s]$ and $\mathrm{E}[s^2]$, for the details, refer to the Section \ref{sec:app:comp},
\begin{equation}
\begin{aligned}\small
    \mathrm{E}[s] &= s_c + \frac{2 h}{3 t} \cdot \mathbf{d}^T \mathbf{n}_c, \\
    \mathrm{E}[s^2] &= s_c^2 + \frac{t^2 h^2 \cdot (\mathbf{d}^T \mathbf{n}_c)^2 +4 t \cdot h^2 s_c \cdot \mathbf{d}^T \mathbf{n}_c}{3 t^2 + h^2},
\end{aligned}
\end{equation}
and thus,
\begin{equation}\small
    \mathrm{E}[(\Phi(\mathbf{x}) - \mu_k) ^ 2] = \mathrm{E}[s^2] - 2 \cdot \mu_k \mathrm{E}[s] + \mu_k^2.
\end{equation}
Replacing the SDF value with the expectation in Eq.~\ref{equ:gmm_density} gives the integrated Gaussian mixture SDF-to-density function:
\begin{equation}\small
    \sigma(\mathbf{x}) = \sum_{k=0}^K \alpha_k \cdot \exp(-\frac{\mathrm{E}[(\Phi(\mathbf{x}) - \mu_k)^2]}{\beta_k^2}),
\end{equation}
Note the proposed integrated SDF can also be used in other SDF-to-density functions~\cite{VolSDF21,NeuS21} $\Psi(\mathrm{E}[\Phi(\mathbf{x})])$. 

\subsection{Training}
\label{sec:train}

Our framework consists of three modules with learnable parameters: i) the SDF network $\Phi(\mathbf{x})$, ii) the implicit renderer $\textbf{C} = M(\hat{\mathbf{x}}, \hat{\mathbf{n}}, \mathbf{d}, f)$, and iii) the SDF-to-density function $\Psi(s)$. 
We train our network by randomly sampling a set of pixels on each training image and minimize the sum of the overall loss computed on each pixel:
\begin{equation}\small
    \mathcal{L} = \mathcal{L}_{rgb} + w_{mask} \mathcal{L}_{mask} + w_{E} \mathcal{L}_{E} + w_{Sp} \mathcal{L}_{Sp},
\end{equation}
where $w_{mask}$, $w_{E}$, and $w_{Sp}$ are weights that balance each loss term.

\noindent \textbf{Mask Loss $\mathcal{L}_{mask}$.} We use a state-of-the-art human matting technique~\cite{rvm21} to extract a saliency of the foreground instance, which provides an object likelihood per pixel. We threshold the saliency with $thr_s=0.002$ and $thr_h=0.9$ to generate the mask ground truth $O_h$ and $O_s$ for both SDF $\Phi_h(\mathbf{x})$ and $\Phi_s(\mathbf{x})$. Then a binary cross-entropy (BCE) loss is used to train the SDF network,
\begin{equation}\small
    \mathcal{L}_{mask} = \mathrm{BCE}(\Phi_h(\mathbf{\hat{x}}), O_h) + \mathrm{BCE}(\Phi_s(\mathbf{\hat{x}}), O_s).
\end{equation}

\noindent \textbf{Photometric Loss $\mathcal{L}_{rgb}$.} We compute the L1 loss between the constructed RGB value and the ground truth value:
\begin{equation}
    \mathcal{L}_{rgb} = \sum_{p \in \mathcal{M}} \lVert C_p - \hat{C}_p \rVert,
\end{equation}
where $\mathcal{M}$ is the mask of foreground regions, $\hat{C}_p$ is the groundturth color on pixels $p$.

\noindent \textbf{Specularity Loss $\mathcal{L}_{Sp}$.} We introduce a new loss to regularize the spurious specularity appearing during novel view synthesis, which could cause color drifting to white or black. This loss penalizes sudden changes in the derivatives of appearance w.r.t. viewing direction, that is,
\begin{equation}\small
    \mathcal{L}_{Sp} = \lVert \frac{\partial M(\hat{\mathbf{x}}, \hat{\mathbf{n}}, \mathbf{d}, f)}{\partial \mathbf{d}} \rVert_2.
\end{equation}

\noindent \textbf{Eikonal Loss $\mathcal{L}_{E}$.} We use the Eikonal regularization on both SDFs~\cite{gropp2020implicit}, \ie, 
\begin{equation}\small
    \mathcal{L}_{E} = \mathrm{E}_\mathbf{x}[(\lVert \nabla \Phi_h(\mathbf{x}) \rVert - 1)^2] + \mathrm{E}_\mathbf{x}[(\lVert \nabla \Phi_s(\mathbf{x}) \rVert - 1)^2].
\end{equation}
where $\mathbf{x}$ is uniformly distributed inside the scene.

\noindent \textbf{Additional Details.} We follow IDR~\cite{IDR20} to pass through gradients toward the traced surface points. Note that different from IDR, in HISR, the differentiable surface points get gradients not only from the implicit renderer, but also from sampling locations in the volume rendering. We also improve the surface tracing process, for additional details (Section \ref{sec:app:surf}).

%% file: Experiment.tex
\begin{table*}[ht!]
\begin{minipage}[]{0.99\linewidth}
\centering
\includegraphics[width=\textwidth]{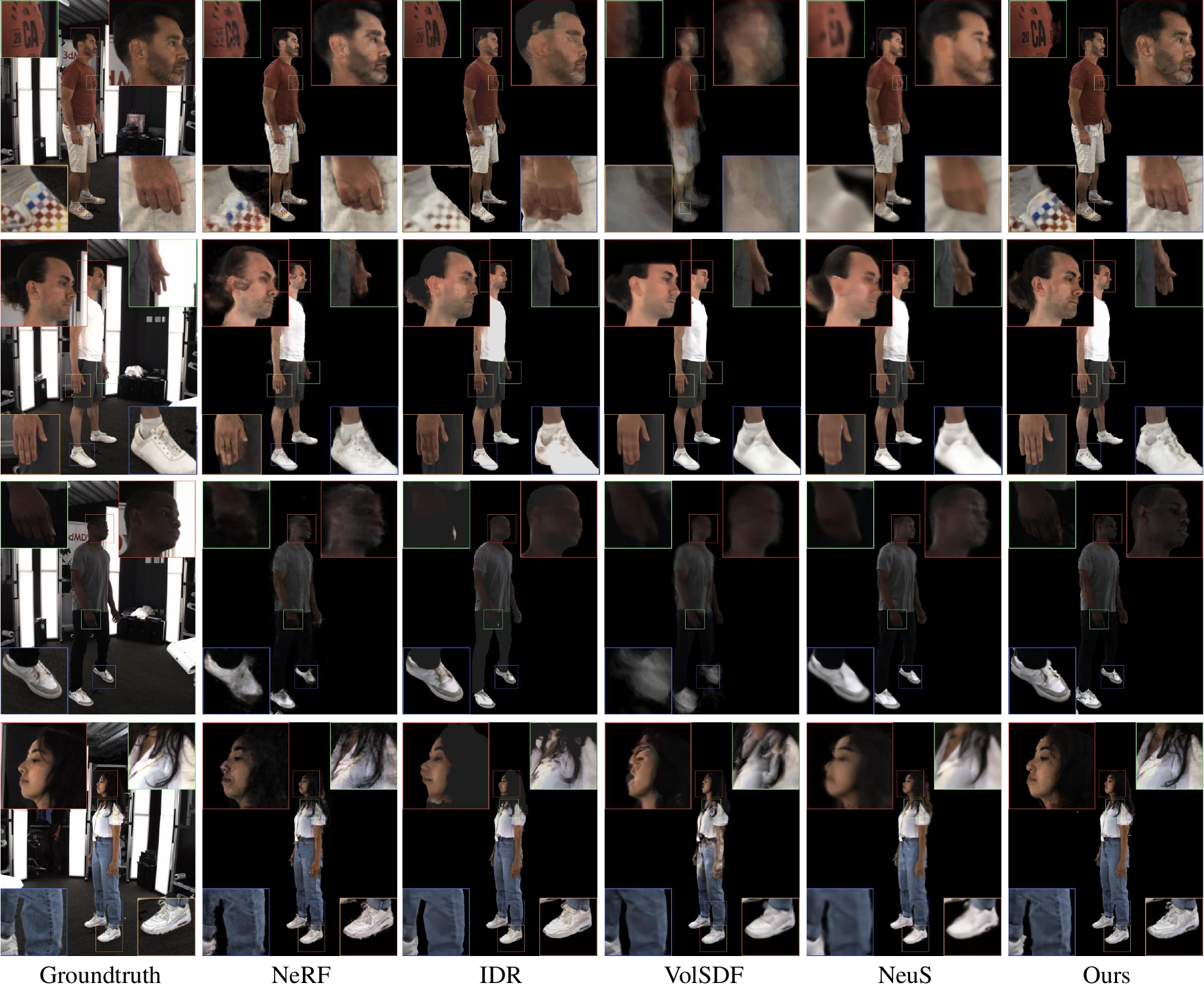}
\captionof{figure}{\textit{Qualitative comparisons with state-of-the-art approaches on novel view synthesis on WIDC dataset}. To protect their privacy, we mask out part of people's faces.}
\label{fig:widc_nvs}
\end{minipage}
\begin{minipage}[]{0.99\linewidth}
\renewcommand\arraystretch{1.1}
\resizebox{\linewidth}{!}{
\centering
\begin{tabular}{l|rrrrrrrrrrrr|r}
\hline\thickhline
PSNR $\uparrow$ & \multicolumn{1}{c}{S1.1} & \multicolumn{1}{c}{S1.2} & \multicolumn{1}{c}{S1.3} & \multicolumn{1}{c}{S2} & \multicolumn{1}{c}{S3} & \multicolumn{1}{c}{S4} & \multicolumn{1}{c}{S5.1} & \multicolumn{1}{c}{S5.2} & \multicolumn{1}{c}{S6} & \multicolumn{1}{c}{S7} & \multicolumn{1}{c}{S8.1} & \multicolumn{1}{c}{S8.2} & \multicolumn{1}{|c}{Avg.} \\
\hline
Nerf      & 28.62                     & 31.65                     & 29.62                     & 24.72                     & 29.33                     & 32.67                     & 26.42                     & 26.76                     & 26.06                     & 23.39                     & 21.21                     & 27.47                     & 27.33                    \\
IDR       & 27.98                     & 30.73                     & 31.37                     & 24.84                     & 29.33                     & 30.01                     & 26.40                     & 26.39                     & 26.74                     & 27.74                     & 28.55                     & 28.85                     & 28.24                    \\
VolSDF    & 28.01                & 30.37                & 30.28                & 24.52                & 25.94                & 29.79                & 25.75                & 26.28                & 21.92                & 23.44                & 31.14                & 30.51                & 27.33                                  \\
Neus      & 29.05                     & 31.91                     & 32.30                     & 26.17                     & 30.76                     & 32.90                     & 27.54                     & 27.48                     & 27.77                     & 28.86                    & 31.88                     & 32.19                     & 29.90                    \\
GS & 30.39 & 33.19 & 33.58 & 27.68 & 33.61 & 34.89 & 30.49 & 30.24 & 30.23 & 30.21 & 33.60 & 34.15 & 31.85 \\
Ours      & \textbf{31.51} & \textbf{34.41} & \textbf{34.07} & \textbf{28.57} & \textbf{33.76} & \textbf{35.14} & \textbf{30.62} & \textbf{31.34} & \textbf{30.51} & \textbf{31.16} & \textbf{34.49} & \textbf{34.48} & \textbf{32.49}                   \\
\bottomrule
\end{tabular}
}
\caption{\textit{Quantitative results and comparisons of PSNR between novel view synthesis and ground truth captures on WIDC dataset}. Best scores are in \textbf{bold}.}
\label{tab:quan_comp}
\end{minipage}
\end{table*}

\section{Experiments}
\label{sec:exp}


\subsection{Implementation Details}

We implement our framework using PyTorch. Specifically, we use the Adam optimizer~\cite{kingma2014adam} with the learning rate $lr=0.005$ and train for 6000 epochs per scene. At each training step, we randomly sample 7200 pixels with 80\% inside the saliency. For volume rendering, we adaptively sample at most 400K points among all viewing rays in the translucent regions at each rendering step.
We use a NVIDIA Tesla V100 GPU for training and it takes \mytilde$30 \text{h}$ to train on each instance, and $110$s to render each image in the original 2K resolution. As a comparison, NeRF takes $204$s, IDR $68$s, NeuS $380$s and VolSDF $670$s.

\begin{figure}
    \centering
    \includegraphics[width=0.46\textwidth]{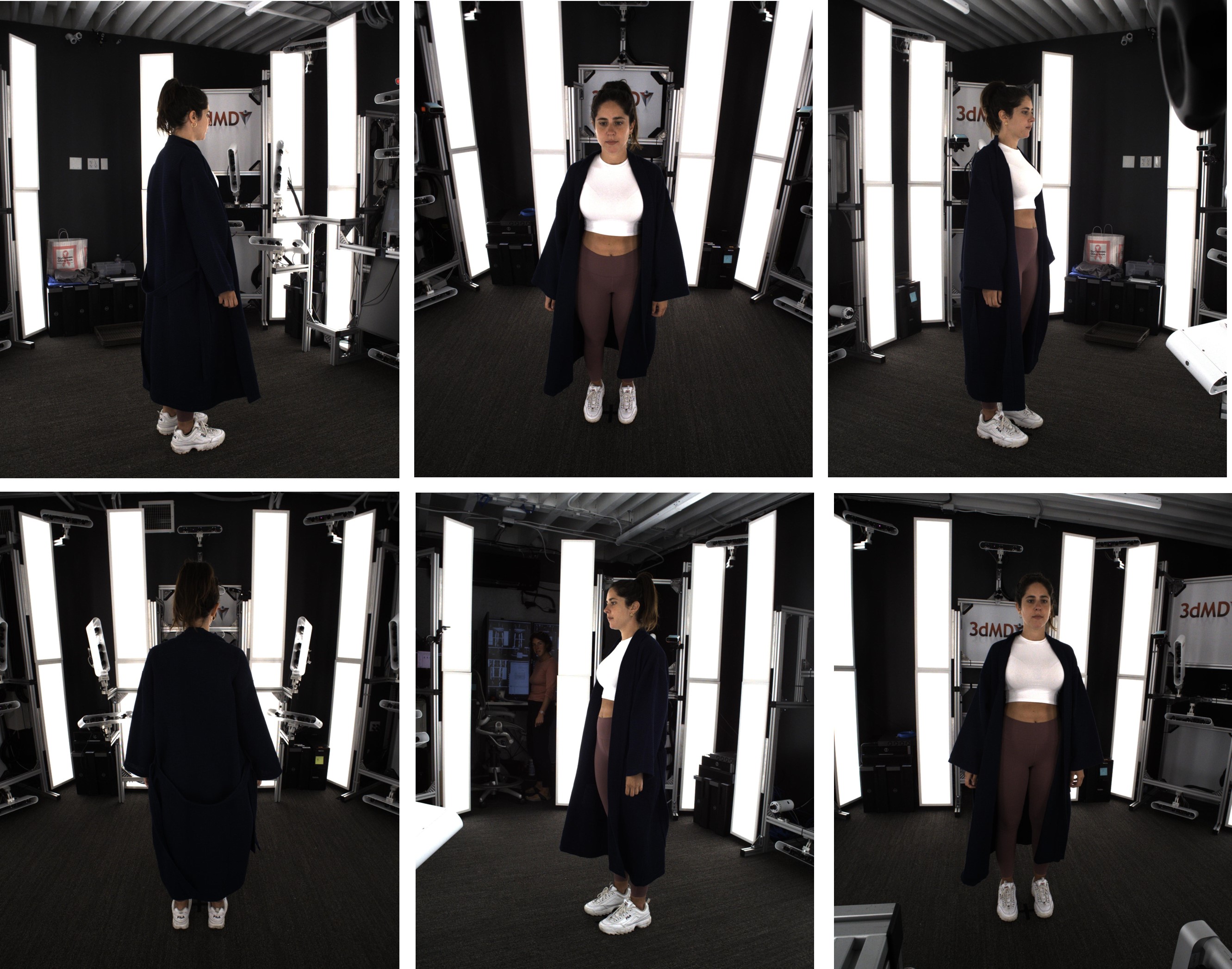}
    \caption{Examples of testing images in WIDC dataset.}
    \label{fig:testing_widc}
\end{figure}

\subsection{Datasets}

We evaluate the effectiveness of our approach on three diverse datasets, encompassing real and synthetic human data, as well as a separate dataset for objects. This comprehensive evaluation allows us to showcase the broad applicability and versatility of our approach across different applications.

\noindent \textbf{WIDC} dataset contains sequences of real humans in motion captured with a 3dMD full-body scanner. The 3dMD scanner comprises 32 to 35 calibrated high-resolution RGB cameras ($2048 \times 2448$) that capture a human in motion performing various actions and facial expressions and output a reconstructed 3D geometry and texture per frame. These scans can be noisy but capture facial expressions and fine-level details like cloth wrinkles. For each instance, we use 26 cameras for training, which focus on different body parts, and the other 6 to 9 cameras for evaluation (as shown in Figure \ref{fig:testing_widc}), which capture the entire human. Examples of training images are provided in the \ref{fig:widc_example}.

\begin{table*}[ht!]
\begin{minipage}[]{0.99\linewidth}
\centering
\includegraphics[width=\textwidth]{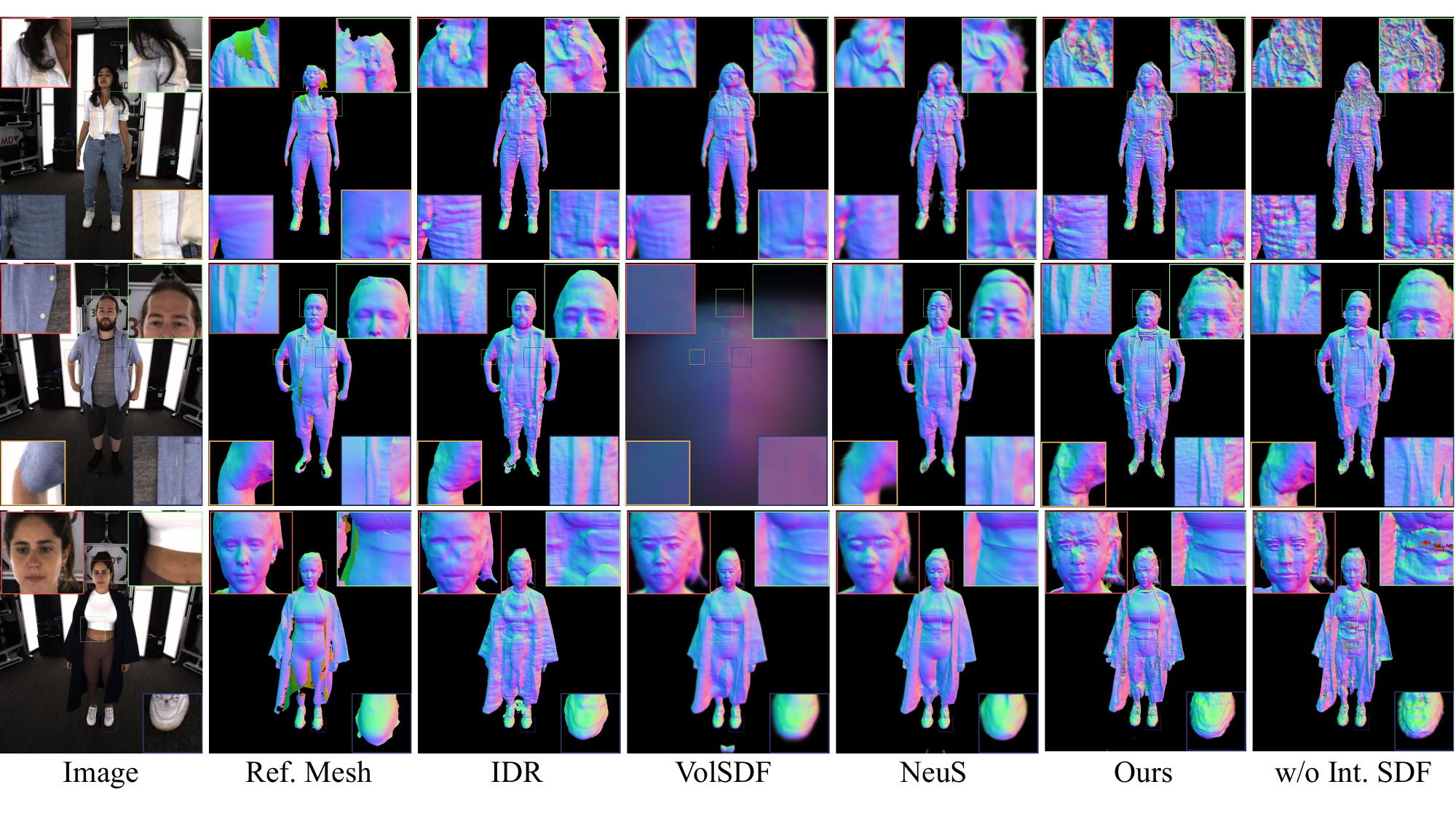}
\captionof{figure}{\textit{Qualitative comparisons with state-of-the-art approaches on geometry reconstruction.} To show the details, we visualize both ours and baseline method with reconstructed surface normals. \textit{w/o} Int. SDF is the ablation experiment for ours without the proposed integrated SDF.}
\label{fig:widc_normal}
\end{minipage}

\begin{minipage}[]{0.99\linewidth}
\renewcommand\arraystretch{1.1}
\resizebox{\linewidth}{!}{
\centering
\begin{tabular}{l|cccccccccccc|c}
\hline\thickhline
CD$^\ast$ $\downarrow$ & \multicolumn{1}{c}{S1.1} & \multicolumn{1}{c}{S1.2} & \multicolumn{1}{c}{S1.3} & \multicolumn{1}{c}{S2} & \multicolumn{1}{c}{S3} & \multicolumn{1}{c}{S4} & \multicolumn{1}{c}{S5.1} & \multicolumn{1}{c}{S5.2} & \multicolumn{1}{c}{S6} & \multicolumn{1}{c}{S7} & \multicolumn{1}{c}{S8.1} & \multicolumn{1}{c}{S8.2} & \multicolumn{1}{|c}{Avg.}         \\
       \hline
Nerf   & 10.94                       & 7.39                         & 10.76                        &     --      & --                  & 40.10                       & 12.80                        & 11.81                       & 67.09                       &      --     & 68.56                       &    --                         &       --                      \\
IDR    & 4.14                        & 5.56                         & 2.57                         & 5.66                        & 4.44                        & 83.61                       & 3.85                         & 7.62                        & 18.71                       & \bf{3.68}                        & 36.36                       & 25.68 & 16.82 \\
VolSDF & 4.10                 & 5.71                 & \textbf{1.05}                 & 8.09                 & 36.80                & 96.88                & 3.45                 & 8.79                 & 29.63                & 99.93                & 22.44                & 20.06                & 28.08   \\
Neus   & 2.92                        & 3.26                         & 1.43                         & 5.05                        & \textbf{2.43 }                       & 26.14                       & 5.16                         & 6.94                        & 5.02                        & 7.07                        & \textbf{11.09}                       &  \textbf{14.60 }&  7.59  \\
Ours   & \bf{1.39}                        & \bf{2.06}                         & 1.83                    & \textbf{4.81}                        & 6.30                        & \bf{7.87}                        & \bf{3.31}                         & \bf{6.16}                        & \bf{3.86}                        & 4.46                        & 11.98                       &  17.79 &  \bf{5.99}  \\
\hline\thickhline
\end{tabular}
}
\caption{\textit{Quantitative results and comparisons of Chamfer Distance (CD) between novel view synthesis and reference mesh reconstruction in WIDC dataset}. Best scores are in \textbf{bold}.}
\label{tab:quan_comp_geo}
\end{minipage}
\end{table*}

\noindent \textbf{SynHuman} dataset is generated by rendering a high-resolution animated 3D human model wearing synthetic clothes, including a simulated t-shirt and pants. Additionally, the dataset includes a hair groom with realistic hair strands, presenting a challenging test environment for our proposed approach. For training, we render images from 24 different cameras, while evaluation is conducted using images rendered from 6 distinct cameras for each instance. In total, we evaluate our approach on three instances within the dataset, allowing for a comprehensive assessment of its performance.

\noindent \textbf{THUman} dataset \cite{deepcloth_su2022} consists 128 viewing cameras from 4 scenes, with $\sim10$ invalid cameras for each scene (Figure \ref{fig:train_thuman}). Although we attempted to align the camera calibration as provided by the authors, we encountered an issue with slight displacement in the translation of the instance. This discrepancy may stem from the PyTorch3D camera system we utilized, which does not account for lens distortion in the capture system. To mitigate this, we optimized the translation during the training phase. Specifically, we employed an Adam optimizer to adjust the object's translation for each camera, using a learning rate of $1e-4$. While we successfully trained our model on the THUman dataset, quantitative evaluations remain challenging due to the calibration issue. Also, due to the same reason, we are not able to provide comparisons with baselines on this dataset.

\noindent \textbf{DTU} dataset \cite{aanaes2016large} comprises multi-view images (49 or 64 views) of various objects, captured using a light stage setup. This dataset provides a ground truth point cloud that serves as the basis for evaluating geometry reconstruction. In our work, we follow the settings established in previous research~\cite{IDR20}. Specifically, we utilize a subset of 15 scenes from the dataset following previous works.

\subsection{Baselines and Evaluation Metrics}

\textbf{Baselines} consist of NeRF~\cite{nerf20}, IDR~\cite{IDR20}, VolSDF~\cite{VolSDF21}, NeuS~\cite{NeuS21}, and Gaussian Splatting (GS)~\cite{kerbl20233d}. For each baseline, we adapt the ray sampling and scene boundary to ensure the instance is placed inside the sampled range. Each baseline is trained for 6000 epochs on each scene. For NeuS and VolSDF, the mask loss proposed in NeuS is used during training. We observe NeRF and VolSDF may fail to converge and might generate empty images. If this happens, we retrain the approach until convergence and stop after three unsuccessful tries.

\begin{table*}
\begin{minipage}[]{0.21\linewidth}
\renewcommand\arraystretch{1.1}
\resizebox{\linewidth}{!}{
\begin{tabular}{@{}r|cc@{}}
\hline\thickhline
        & PSNR $\uparrow$ & CD $\downarrow$ \\
\hline
IDR &    31.65                      &  6.85 \\
VolSDF &  32.43                 & 7.10   \\
NeuS &  33.21   & 10.58   \\
Ours &  \bf{35.69}                   & \bf{5.66}                \\
\hline\thickhline
\end{tabular}}
\caption{\textit{Quantitative comparisons on SynHuman dataset}.}
\label{tab:truetony}
\end{minipage}
\hspace{2mm}
\begin{minipage}[]{0.76\linewidth}
\resizebox{\linewidth}{!}{
\includegraphics[width=0.49\textwidth]{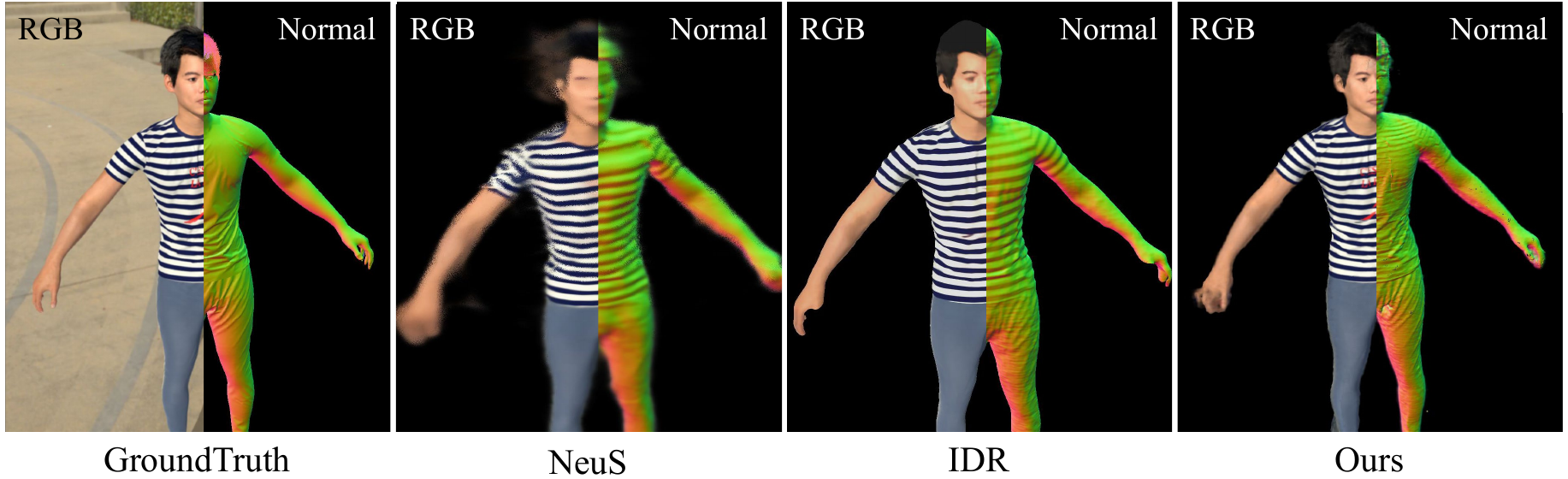}
}
\captionof{figure}{\textit{Qualitative comparisons for reconstructed geometry and novel view synthesis on SynHuman dataset}. }
\label{fig:truetony}
\end{minipage}
\end{table*}

\begin{figure*}
    \centering
    \includegraphics[width=\textwidth]{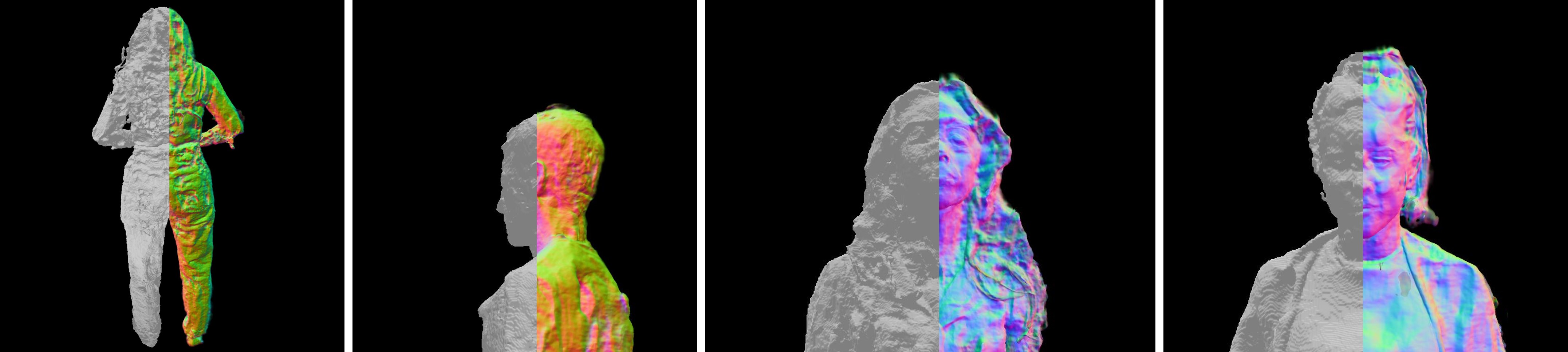}
    \caption{Comparison for reconstructed surface normals and reconstructed meshes. Specifically, we render the reconstructed meshes of Ours on the same pose of the rendered surface normals, and we concatenate them side-by-side.}
    \label{fig:normalvsrec}
\end{figure*}

\begin{table}
\centering
\begin{tabular}{r|cc}
\hline\thickhline
& PSNR $\uparrow$   & CD $^\ast$ $\downarrow$ \\
\hline
with Gaussian Density    & 29.91 & 17.30 \\
with Laplacian Density      & 29.76 & 30.86 \\
w/o integrated SDF    & 30.28 & 20.41 \\
w/o specularity loss  & 25.25 & 23.18 \\
with Translucent region only & 25.23 & --- \\
with Opaque region only & 27.81 & --- \\
full model            & \bf{30.34} & \bf{5.99} \\
\hline\thickhline
\end{tabular}
\caption{\textit{Ablation studies on WIDC dataset.} Best scores are in \textbf{bold}.}
\label{tab:ablation}
\end{table}

\begin{table*}[ht!]
\hspace{4mm}
\begin{minipage}[]{0.96\linewidth}
\centering
\includegraphics[width=\textwidth]{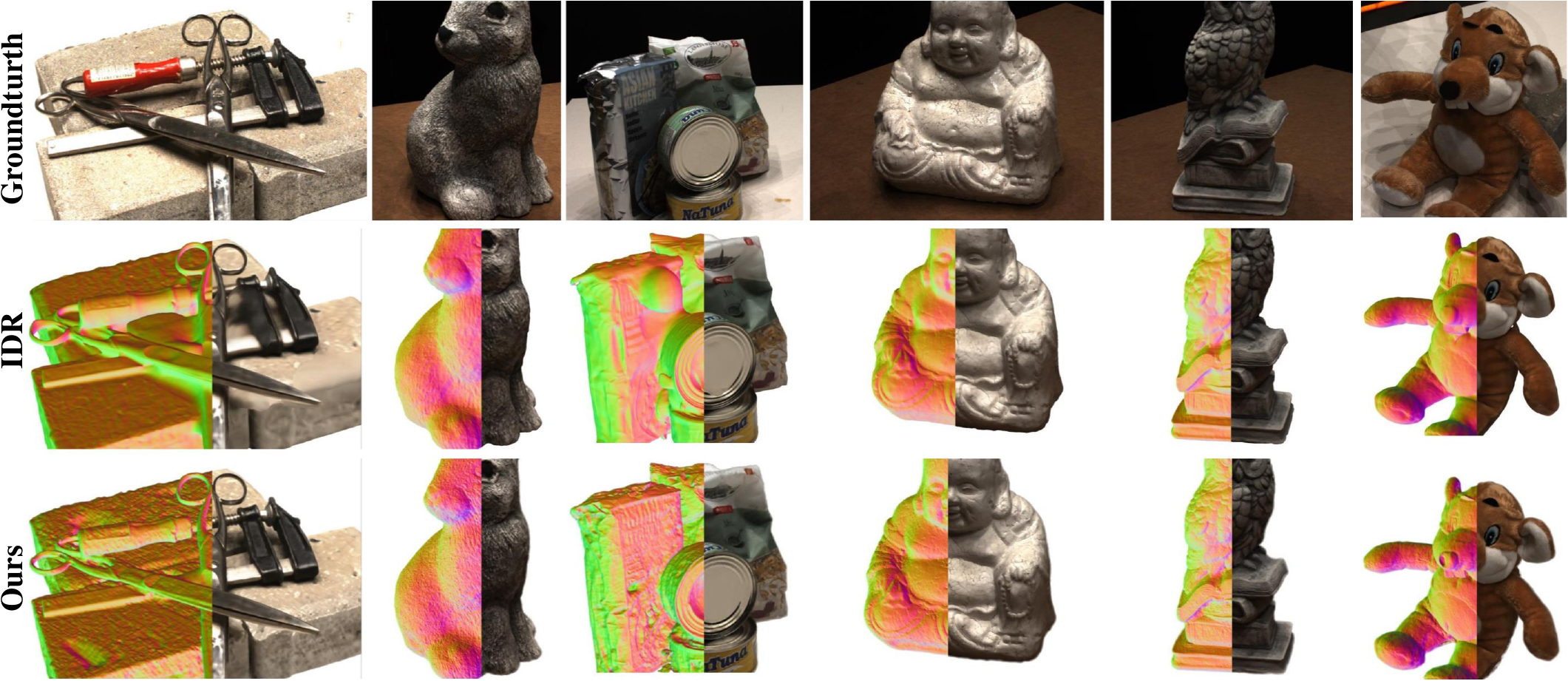}
\captionof{figure}{\textit{Qualitative comparisons for reconstructed geometry and novel view synthesis on DTU Dataset}. }
\label{fig:dtu}
\end{minipage}
\end{table*}

\begin{table}[!ht]
    \centering
    \begin{tabular}{c|cc|cc}
    \hline\thickhline
    \multicolumn{1}{c|}{DTU}        & \multicolumn{2}{c|}{PSNR $\uparrow$}                           & \multicolumn{2}{c}{CD $\downarrow$} \\
    \hline
     Scene ID    & IDR & Ours & IDR              & Ours              \\
    \hline
    24  & 23.55 & 24.39 & 1.24 & 0.90  \\
    37  & 21.08 & 22.53 & 1.22 & 0.56 \\
    40  & 24.62 & 25.66 & 0.87 & 1.08 \\
    55  & 23.64 & 24.07 & 0.39 & 0.33 \\
    63  & 25.47 & 26.47 & 0.54 & 0.39 \\
    65  & 23.26 & 26.55 & 0.82 & 0.54 \\
    69  & 22.36 & 24.31 & 0.42 & 0.49 \\
    83  & 21.97 & 24.42 & 2.22 & 2.17 \\
    97  & 23.16 & 24.18 & 0.92 & 0.51 \\
    105 & 22.97 & 26.73 & 0.99 & 1.00    \\
    106 & 22.17 & 26.04 & 0.51 & 0.84 \\
    110 & 23.07 & 23.86 & 0.97 & 1.36 \\
    114 & 25.04 & 25.81 & 0.25 & 0.43 \\
    118 & 24.18 & 26.25 & 0.54 & 0.49 \\
    122 & 27.42 & 27.30 & 0.63 & 0.44  \\
    \hline
    Average & 23.60      & \bf{25.24}              &  0.84                &   \textbf{0.77}        \\
    \hline\thickhline
    \end{tabular}
    \caption{\textit{Quantitative comparisons on DTU dataset}.}
    \label{tab:dtu}
\end{table}

\noindent \textbf{Evaluations} are conducted for measurement of both novel view synthesis and reconstructed geometries. Specifically, we evaluate predicted novel views via PSNR, which computed via the L2 distance of groundtruth and reconstructed images. For the geometry, we first use the Marching Cubes algorithm~\cite{lorensen1987marching} to extract the surface mesh of the trained SDF. We use single-direction Chamfer Distance (CD$^\ast$) for evaluation, which only computes the L1 distance from each vertex on ground truth mesh to the constructed one. This is because the captured 3D scans (\ie, reference mesh) from 3dMD system may contain missing surfaces, as shown in Figure~\ref{fig:widc_normal}. For DTU datasets, we use the standard evaluation metrics PSNR and CD, following~\cite{IDR20}. We also qualitatively evaluate the reconstructed geometry by visualizing the surface normals.

\subsection{Quantitative Results and Qualitative Comparisons}

\noindent \textbf{WIDC.} 
In Table~\ref{tab:quan_comp} and Table~\ref{tab:quan_comp_geo}, we provide a quantitative comparison between our method and various strong baselines. The results demonstrate that our approach outperforms all works in terms of both RGB reconstruction and reconstructed geometries. It is worth noting that the NeRF baseline may encounter difficulties in reconstructing via Marching Cubes if there are insufficient volumes with densities larger than the threshold value of $thr=50$. In such cases, these entries in the table are denoted as "---".
Figures \ref{fig:widc_nvs} and \ref{fig:widc_normal} show the reconstructed geometries and novel view synthesis. The visualizations clearly showcase that our method excels in capturing fine-level geometric details compared to other reconstruction techniques. It is notable that our approach achieves better performance across the board and accurately reconstructs people with different hair styles and skin tones. 
Furthermore, our method exhibits a high level of rendering fidelity on the boundaries, surpassing approaches that solely rely on volume rendering. 

\noindent \textbf{SynHuman.} 
In addition, we showcase the outcomes on the SynHuman dataset in Figure \ref{fig:truetony} and Table \ref{tab:truetony}. The qualitative results highlight a substantial improvement in cloth reconstruction by our method compared to the baselines. Specifically, our approach yields more precise wrinkle details and lessens texture confusion with the geometries of the cloth. These outcomes demonstrate the exceptional reconstruction quality attained through our methodology.

\noindent \textbf{THUman.} 
Qualitative results demonstrate our approach performs well in general 3D human reconstruction applications (Figure \ref{fig:thuman}). 
From the results, we can observe we obtain high-fidelity reconstruction on both cloth and hair.
Also, the results show our approach is feasible to achieve significantly better reconstruction quality with more multi-view input images, even with inaccurately calibrated cameras.

\noindent \textbf{DTU.} 
Table \ref{tab:dtu} and Figure \ref{fig:dtu} show the reconstruction results on the DTU dataset.
The quantitative results illustrate significant improvements achieved by our method compared to the baseline method, \ie IDR.
Experiments on DTU dataset demonstrate that although HISR is specially designed for reconstruction of 3D humans, it can effectively perform reconstruction of generic objects. 
The visualization of surface normals shows ours more fine-grained details on the reconstructed geometry.
As shown in the first column in Figure \ref{fig:dtu}, ``ours'' achieves significantly better reconstruction of the geometry of the region under the scissors on the top.

\subsection{Ablation Studies}

In Table \ref{tab:ablation}, we provide an ablation study to assess the individual contributions of each proposed module. In particular, we ablate \emph{the Laplace function} and \emph{the single Gaussian function} used in VolSDF and NeuS, which serve as replacements for our proposed Gaussian Mixture density.
The results clearly indicate that these choices significantly reduce the quality of the reconstruction.
Furthermore, we evaluate HISR \emph{without the integrated SDF} module, by relying solely on the SDF value at the center of the conical frustum. 
For this ablation, we also show a qualitative comparison with our full approach in Figure \ref{fig:widc_normal}.
The results highlight the substantial contribution of the integrated SDF in achieving accurate reconstructed geometry.
Then we perform an ablation study on the \emph{specularity loss}, demonstrating its efficacy as a valuable regularizer that benefits both novel view synthesis and reconstruction tasks.
Finally, we also report the novel view synthesis results when \emph{only rendering translucent or opaque regions}, the results demonstrate both translucent or opaque are significantly contribute to the final reconstruction.

%% file: AppendixS.tex
\section{Proof and Computation Details}

In this section, we include proof of the formulations in the main paper.

\subsection{SDF Values in Uniform Parallel Field (Proof of Theorem 1)}
\label{sec:app:proof}
\begin{figure}[h]
    \centering
    \includegraphics[width=0.35\textwidth]{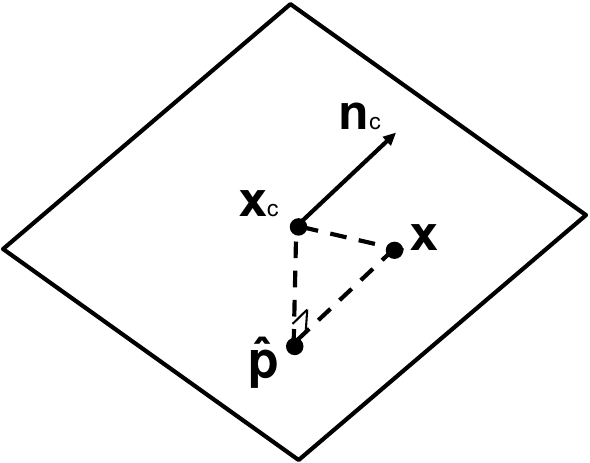}
    \caption{\textit{Illustration of $\mathbf{x}$ and $\mathbf{x}_c$.}}
    \label{fig:theom1}
\end{figure}
Here we prove \textbf{Theorem 1}, which computes the SDF value by the assumption of a uniform parallel field. As Figure \ref{fig:theom1} shows, given the known SDF value $s_c$ at location $\mathbf{x}_c$, we will have a plane $\mathcal{G}$ that $\mathbf{x}_c \in \mathcal{G}$, and the normalized gradient direction $\mathbf{n}_c, \lVert \mathbf{n}_c \rVert=1$ at $\mathbf{x}_c$ is vertical to $\mathcal{G}$ that
\begin{equation}
    \forall \mathbf{p} \in \mathcal{G}, (\mathbf{p} - \mathbf{x}_c) \cdot \mathbf{n}_c = 0.
\end{equation}
To compute the SDF value for any location $\mathbf{x}$ in the field, we first compute the distance from $\mathbf{x}$ to $\mathcal{G}$. Since the distance from a point to a plane is along a line perpendicular to the plane. Hereby we denote the intersection of the perpendicular line and $\mathcal{G}$ as $\mathbf{\hat{p}}$, so that $\mathbf{x} - \mathbf{\hat{p}}$ and $\mathbf{n}_c$ is on the same line. Now, we have
\begin{equation}
\begin{split}
    \lVert \mathbf{x} - \mathbf{\hat{p}} \rVert &= | \frac{ (\mathbf{x} - \mathbf{\hat{p}}) \cdot \mathbf{n}_c}{ \lVert \mathbf{n}_c \rVert} | \\ 
    &= | (\mathbf{x} - \mathbf{\hat{p}}) \cdot \mathbf{n}_c + ( \mathbf{\hat{p}} - \mathbf{x}_c) \cdot \mathbf{n}_c | \\
    &= | \mathbf{n}_c \cdot (\mathbf{x} - \mathbf{x}_c) |.
\end{split}
\end{equation}
Since the $\mathbf{x} - \mathbf{\hat{p}}$ place along the direction of the SDF field, we have $|s - s_{\hat{p}} | = \lVert \mathbf{x} - \mathbf{\hat{p}} \rVert = | \mathbf{n}_c \cdot (\mathbf{x} - \mathbf{x}_c) |$. Note $s_{\hat{p}} = s_c$ since the SDF values are the same on any location of the plane. In the case that $\mathbf{x}$ located along the gradient side of the plane $s - s_{\hat{p}} = s - s_c = \mathbf{n}_c \cdot (\mathbf{x} - \mathbf{x}_c) $, vice versa. Thus, 
\begin{equation}
    \Phi(\mathbf{x}) = s = s_c + \mathbf{n}_c \cdot (\mathbf{x} - \mathbf{x}_c).
    \label{equ:theo1}
\end{equation}

\begin{figure}[t!]
    \centering
    \includegraphics[width=0.47\textwidth]{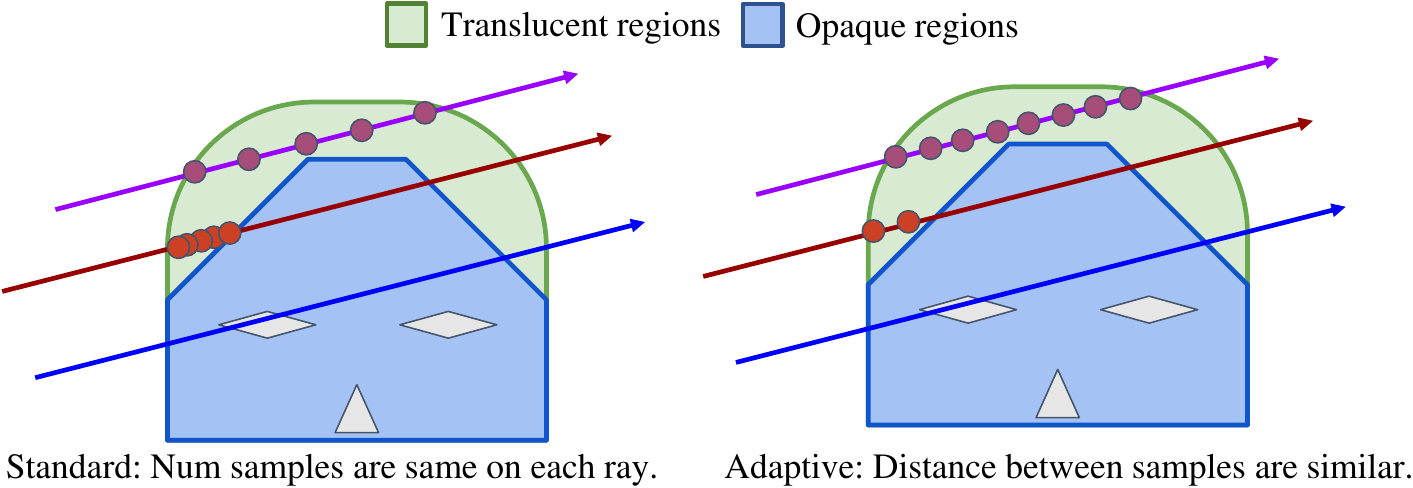}
    \caption{\textit{Illustration of adaptive ray sampling}, which controls the sampling interval on each ray to be similar.}
    \label{fig:adaptive_sample}
\end{figure}

\subsection{Computation Details on Expected SDF Values}
\label{sec:app:comp}
Here we compute the expectation of SDF values inside each conical frustum. The expectation is computed as,
\begin{equation}\small
    \mathrm{E}[s] = \frac{\int \Phi(\mathbf{x}) \cdot \mathrm{F}\left(\mathbf{x}, \mathbf{o}, \mathbf{d}, r, t, h\right) d\mathbf{x}}{\int \mathrm{F}\left(\mathbf{x}, \mathbf{o}, \mathbf{d}, r, t, h\right) d\mathbf{x}},
\end{equation}
where $\mathrm{F}(\mathbf{x}, \cdot)$ denotes the space inside the conical frustum, $\Phi(\mathbf{x})$ is the SDF function.
Based on Equation \ref{equ:theo1}, we have,
\begin{equation}
\begin{split}
    \mathrm{E}[s] &= \frac{\int (s_c + \mathbf{n}_c \cdot (\mathbf{x} - \mathbf{x}_c) ) \cdot \mathrm{F}\left(\mathbf{x}, \mathbf{o}, \mathbf{d}, r, t, h\right) d\mathbf{x}}{\int \mathrm{F}\left(\mathbf{x}, \mathbf{o}, \mathbf{d}, r, t, h\right) d\mathbf{x}} \\
    &= s_c + \frac{\int  \mathbf{n}_c \cdot (\mathbf{x} - \mathbf{x}_c) \cdot \mathrm{F}\left(\mathbf{x}, \mathbf{o}, \mathbf{d}, r, t, h\right) d\mathbf{x}}{\int \mathrm{F}\left(\mathbf{x}, \mathbf{o}, \mathbf{d}, r, t, h\right) d\mathbf{x}} \\
    &= s_c \\ 
    &+ \frac{\int_{-h}^h \int_{0}^{r \frac{t+a}{t}} \int_{0}^{2\pi} (\hat{n}_x \gamma \sin(\theta) + \hat{n}_y \gamma \sin(\theta) + \hat{n}_z a) \cdot r \mathrm{d} \theta \mathrm{d} \gamma \mathrm{d} a}{\int_{-h}^h \int_{0}^{r \frac{t+a}{t}} \int_{0}^{2\pi} r \mathrm{d} \theta \mathrm{d} \gamma \mathrm{d} a} \\
    &= s_c + \frac{2 h t}{ 3 t^2 + h^2 } \hat{n}_z, 
\end{split}
\end{equation}
where $t$ is the distance for the sample from ray origin $\mathbf{o}$, $2h$ is the height of the conical frustum, $\{ \hat{n}_x, \hat{n}_y, \hat{n}_z \}$ is the rotated normal, and $ \hat{n}_z = \frac{\mathbf{d}^T \mathbf{n}_c}{ \lVert \mathbf{n}_c \rVert \lVert \mathbf{d} \rVert} = \mathbf{d}^T \mathbf{n}_c$, where $\mathbf{d}$ is the viewing ray direction. Specifically, when we compute the range of integral, we rotate the world coordinate to let z-axis be located along the viewing ray. 
Similarly, we compute $\mathrm{E} [ s^2 ]$ via,

\begin{figure*}
    \centering
    \includegraphics[width=\textwidth]{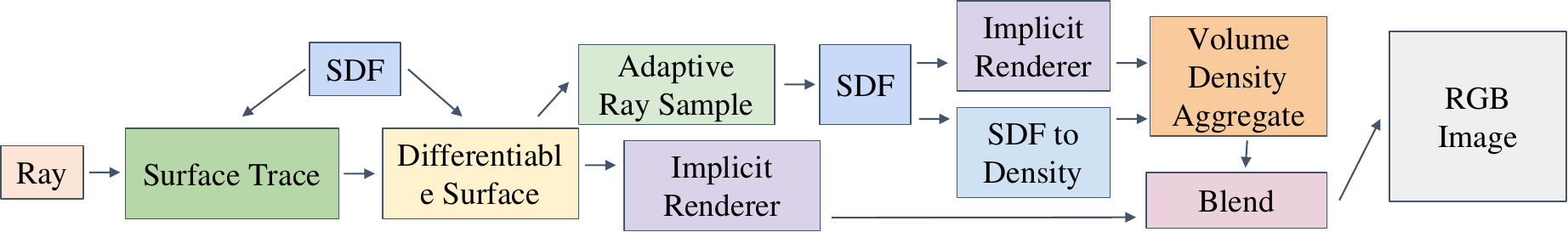}
    \caption{Rendering pipeline for HISR. We take the viewing ray as input, and synthesis the image via blending surface rendering and volume rendering results.}
    \label{fig:render_pipeline}
\end{figure*}

\begin{equation}
\begin{split}
    \mathrm{E}[s^2] &= \frac{\int (s_c + \mathbf{n}_c \cdot (\mathbf{x} - \mathbf{x}_c) )^2 \cdot \mathrm{F}\left(\mathbf{x}, \mathbf{o}, \mathbf{d}, r, t, h\right) d\mathbf{x}}{\int \mathrm{F}\left(\mathbf{x}, \mathbf{o}, \mathbf{d}, r, t, h\right) d\mathbf{x}} \\
    &= s_c^2 + \frac{t^2 h^2 \cdot (\hat{n}_z )^2 +4 t \cdot h^2 s_c \cdot \hat{n}_z }{3 t^2 + h^2}.
\end{split}
\end{equation}
During implementation, since $t \gg h$, we can simplify it as
\begin{equation}
    \mathrm{E}[s^2] =s_c^2 + \frac{h^2 \cdot (\mathbf{d}^T \mathbf{n}_c)^2}{3} + \frac{4 h^2 s_c \cdot \mathbf{d}^T \mathbf{n}_c}{3 t},
\end{equation}
\begin{equation}
    \mathrm{E}[s] = s_c + \frac{2 h }{ 3 t} \mathbf{d}^T \mathbf{n}_c.
\end{equation}

\section{Architecture}
\label{sec:app:arch}

In this section, we include more details and form regarding to our proposed framework.

\begin{figure}[t!]
    \centering
    \includegraphics[width=0.35\textwidth]{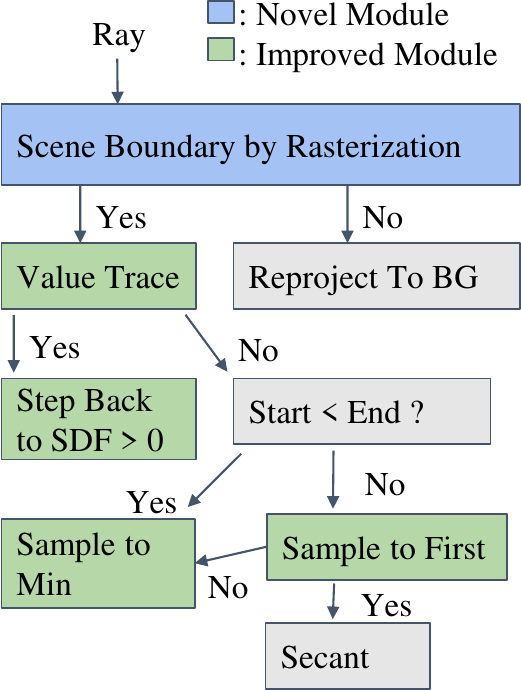}
    \caption{\textit{The surface tracing process} determines the intersection of viewing rays and implicit surfaces. To improve the process, we introduce a novel module while improving the other modules used in previous works.}
    \label{fig:ray_trace}
\end{figure}

\subsection{Pipeline}
\begin{figure*}[t!]
     \centering
     \includegraphics[width=\textwidth]{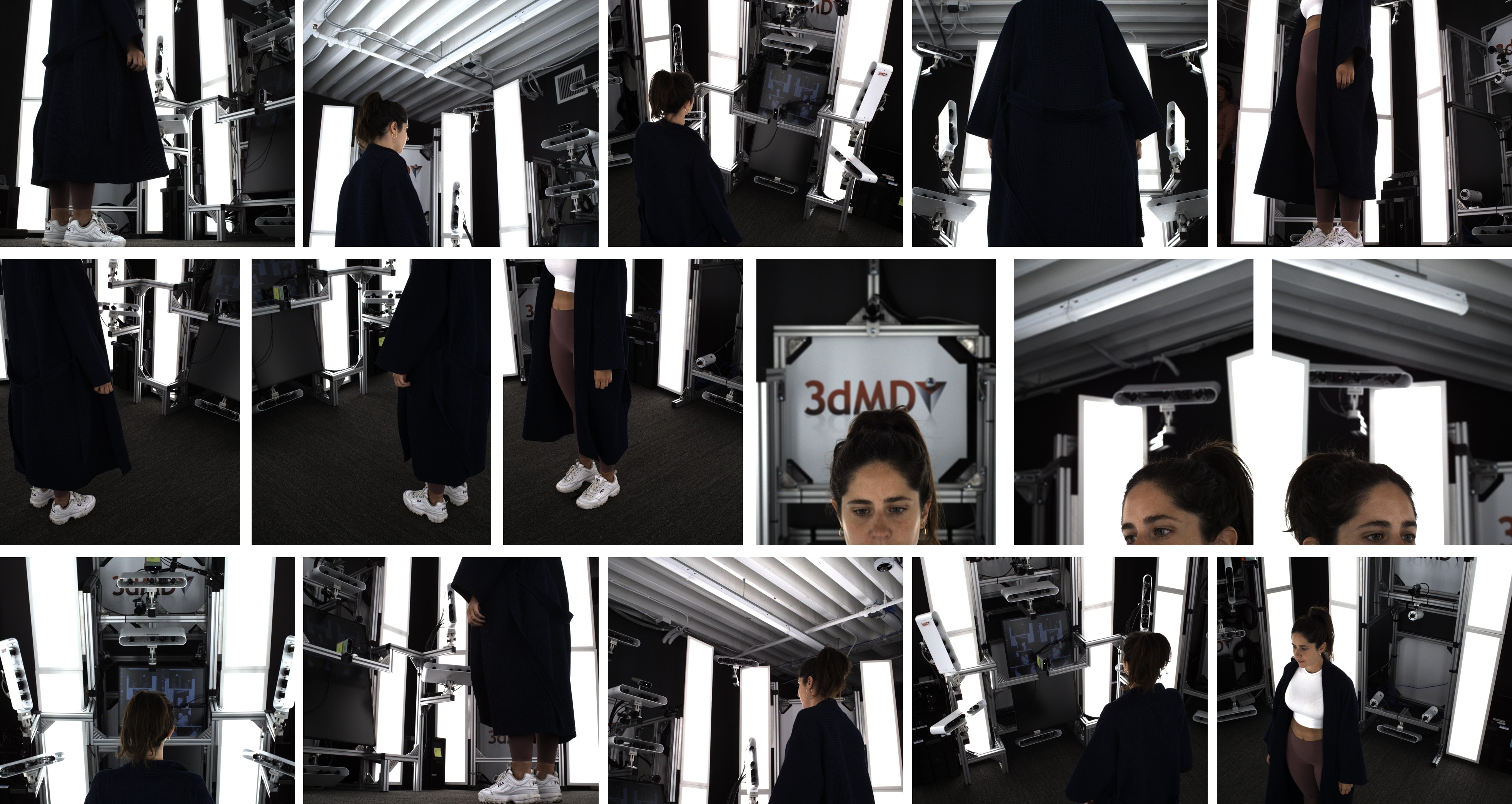}
    \caption{Example of WIDC training images (S8.1).}
    \label{fig:widc_example}
\end{figure*}

Figure \ref{fig:render_pipeline} shows the forward pipeline for HISR in the training stage. Given the viewing ray, we first conduct the surface tracing to indicate 1) whether the ray intersects with hard surface $\mathcal{P}_h$ and soft surface $\mathcal{P}_s$, and 2) if intersects, $\hat{t}_h$, $\hat{t}_s$, $\ddot{t}_s$. Then we compute the differentiable surface point $\mathbf{\hat{x}}$ following IDR,
\begin{equation}
    \mathbf{\hat{x}} = \mathbf{o} + t \mathbf{d} - \frac{\mathbf{d}}{\nabla \Phi(\mathbf{o} + t \mathbf{d})  \cdot \mathbf{d}} \cdot \Phi(\mathbf{o} + t \mathbf{d}),
\end{equation}
where $\mathbf{d}$ is the direction of the viewing ray, $\nabla$ computes the divergence.

Then we conduct surface rendering on the hard surface following IDR,
\begin{equation}
\textbf{C}_h = M(\mathbf{\hat{x}}_h, \hat{\mathbf{n}}_h, \mathbf{d}, f),
\end{equation}
where $M$ is the implicit render.

As discussed in Section 3.1 Hybrid Rendering, we determine the sampling range for the soft render based on the cases of surface tracing. Specifically, we sample from $t=\hat{t}_s$ to $t=\ddot{t}_s$ for case 3, and from $t=\hat{t}_s$ to $t=\hat{t}_h$ for case 4. Each sample is computed via the adaptive sampling strategy as described in Section \ref{sec:ada_sample}. For each sample $\mathbf{x}_k$, we compute the viewing color,
\begin{equation}
\textbf{C}_k = M(\mathbf{\hat{x}}_k, \hat{\mathbf{n}}_k, \mathbf{d}, f),
\end{equation}
and the volume density
\begin{equation}
    \sigma_k = \Psi( \Phi (\mathbf{x}_k) ).
\end{equation}
Then the soft viewing color is computed via the volume rendering formulation,
\begin{equation}
\begin{split}
\mathbf{C}_s=\sum_k T_k\left(1-\exp \left(-\sigma_k\left(t_{k+1}-t_k\right)\right)\right) \mathbf{c}_k, \\
\text { with }  T_k=\exp \left(-\sum_{k^{\prime}<k} \sigma_{k^{\prime}}\left(t_{k^{\prime}+1}-t_{k^{\prime}}\right)\right).
\end{split}
\label{equ:volume_rendering_app}
\end{equation}
Finally, we blend the $\mathbf{C}_h$ and $\mathbf{C}_s$ as discussed in the main text Section 3.1 Hybrid Rendering.

\begin{figure*}
    \centering
    \includegraphics[width=\textwidth]{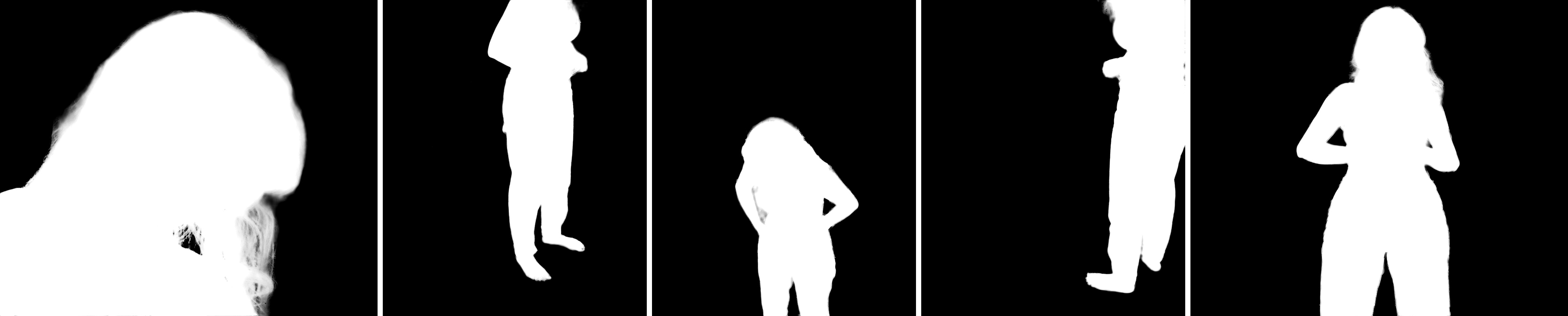}
    \caption{\textit{Example of matting mask obtained by Robust Video Matting.}}
    \label{fig:matting}
\end{figure*}

\subsection{Network Architecture}
We follow \cite{VolSDF21, IDR20, NeuS21} to construct the network architecture of our SDF network and the implicit renderer, but slightly reduce parameters to be 6 layers with 256 channels for the SDF and 6 layers with 256 channels for the implicit renderer. Following \cite{nerf20, IDR20} we use the positional encode to encode both point location $\mathbf{x}$, viewing direction $\mathbf{d}$, and surface normal $\mathbf{n}$. Then we concatenate them as the input of the implicit renderer $M$. The level of frequency for SDF is $\mathbf{x}$: 10, for implicit renderer is $\mathbf{n}$: 4, $\mathbf{d}$: 4.

\subsection{Surface Tracing}
\label{sec:app:surf}
We study and improve the surface tracing process as Figure \ref{fig:ray_trace} shows. Specifically, given a viewing ray $\mathbf{o} + t \mathbf{d}$, we first compute its intersection with a pre-defined outer bounding box mesh. This process is conducted using a rasterizer which gives the first and last intersection distance of the ray given the mesh. Then following IDR, we conduct an SDF value-based trace, which steps forward as the SDF value at the current location for the forward direction, and step back on the backward ray direction. We simultaneously handle the searched points that incorrectly go across the surface and into the object. Such a process is conducted via binary search between the previous distance and current distance until at a location that gives SDF value $s > 0$. To further make the search more efficient, we introduce a learnable step size in the binary search. For those unconvergent rays, we use adaptive sampling to find either the first location the SDF value $s < 0$, or the minimum SDF value on the whole ray. Our implementation of surface tracing achieve same results compared to the previous one used in IDR, and achieve 1.85x times speed up.

\begin{table*}[!t]
    \centering
\renewcommand\arraystretch{1.0}
\small
\begin{tabular}{l|p{8.7mm}p{8.7mm}p{8.7mm}p{8.7mm}p{8.7mm}p{8.7mm}p{8.7mm}p{8.7mm}p{8.7mm}p{8.7mm}p{8.7mm}p{8.7mm}|p{8.7mm}}
\hline\thickhline
    PSNR & \multicolumn{1}{c}{S1.1} & \multicolumn{1}{c}{S1.2} & \multicolumn{1}{c}{S1.3} & \multicolumn{1}{c}{S2} & \multicolumn{1}{c}{S3} & \multicolumn{1}{c}{S4} & \multicolumn{1}{c}{S5.1} & \multicolumn{1}{c}{S5.2} & \multicolumn{1}{c}{S6} & \multicolumn{1}{c}{S7} & \multicolumn{1}{c}{S8.1} & \multicolumn{1}{c}{S8.2} & \multicolumn{1}{|c}{Avg.} \\
    \hline
    NeRF & $28.62 \newline \pm  0.84$  & $31.65 \newline \pm 2.05$  & $29.62 \newline \pm 1.90$  & $24.72 \newline \pm 1.10$  & $29.33 \newline \pm 1.76$  & $32.67 \newline \pm 0.69$  & $26.42 \newline \pm 1.04$  & $26.76 \newline \pm 1.87$  & $26.06 \newline \pm 1.79$  & $23.39 \newline \pm 1.49$  & $21.21 \newline \pm 1.51$  & $27.47 \newline \pm 1.76$  &   $28.73 \newline \pm 1.48$ \\
    IDR & $27.98 \newline \pm 0.92$  & $30.73 \newline \pm 1.03$  & $31.37 \newline \pm 1.10$  & $24.84 \newline \pm 0.98$  & $29.33 \newline \pm 1.31$  & $30.01 \newline \pm 0.76$  & $26.40 \newline \pm 1.28$  & $26.39 \newline \pm 1.41$  & $26.74 \newline \pm 2.03$  & $27.74 \newline \pm 1.82$  & $28.55 \newline \pm 0.50$  & $28.85 \newline \pm 0.39$  &   $28.24 \newline \pm 1.13$ \\
    VolSDF & $28.01 \newline \pm 1.70$  & $30.28 \newline \pm 2.53$  & $30.37 \newline \pm 2.22$  & $24.52 \newline \pm 1.90$  & $25.94 \newline \pm 3.98$  & $29.79 \newline \pm 2.18$  & $25.75 \newline \pm 1.89$  & $26.28 \newline \pm 2.53$  & $21.92 \newline \pm 3.01$  & $23.32 \newline \pm 2.38$  & $31.14 \newline \pm 1.85$  & $30.51 \newline \pm 1.82$  &   $28.72 \newline \pm 2.33$ \\
    NeuS & $28.01 \newline \pm 0.98$  & $31.91 \newline \pm 1.62$  & $32.30 \newline \pm 1.69$  & $26.17 \newline \pm 1.20$  & $30.76 \newline \pm 1.08$  & $32.90 \newline \pm 1.33$  & $27.74 \newline \pm 1.05$  & $27.48 \newline \pm 1.27$  & $27.77 \newline \pm 2.13$  & $28.86 \newline \pm 1.51$  & $31.88 \newline \pm 1.17$  & $32.19 \newline \pm 1.54$  &   $29.90 \newline \pm 1.31$ \\
    ours  & $31.51 \newline \pm 1.18$  & $34.41 \newline \pm 1.55$  & $34.07 \newline \pm 1.40$  & $28.57 \newline \pm 1.35$  & $33.76 \newline \pm 1.03$  & $35.14 \newline \pm 0.94$  & $30.62 \newline \pm 1.44$  & $31.14 \newline \pm 1.92$  & $30.51 \newline \pm 2.20$  & $31.16 \newline \pm 1.70$  & $34.49 \newline \pm 1.36$  & $34.48 \newline \pm 0.78$  &   $32.49 \newline \pm 1.46$ \\
\hline\thickhline
    \end{tabular}
    \caption{Quantitative results on WIDC dataset for Ours and baselines with error bar.}
    \label{tab:error}
\end{table*}


\begin{figure*}[t!]
    \centering
    \includegraphics[width=\textwidth]{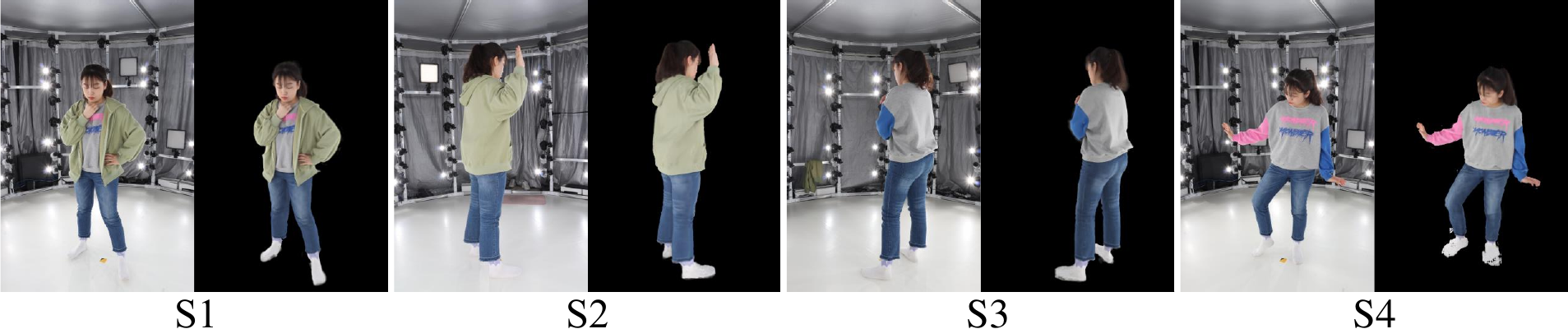}
    \caption{\textit{Qualitative results for novel view synthesis on THUman. For each image pair, we show the comparison between the groundturth image on the left and our result on the right.}}
    \label{fig:thuman}
\end{figure*}

\subsection{Adaptive Ray Sampling}
\label{sec:ada_sample}
As Figure \ref{fig:adaptive_sample} shows, the Adaptive Ray Sampler sample points on a set of rays to achieve similar intervals between each sample. Given the total sampled range,
\begin{equation}
    R = \sum_i r_i,
\end{equation}
where $r_i$ is the sampling range on each ray. We compute the uniform sampling interval as $\delta z = \frac{R}{N}$, where $N$ is the demand total samples. Then the number of samples on each ray is $n_i = \lfloor \frac{r_i}{z} \rfloor$, then we uniformly sample points on each ray via interval $\frac{r_i}{n_i}$. One challenge thing in this process is it is not efficient to compute the sum in volume rendering formulation when the number of samples are different on each ray. To achieve this, we implement a CUDA function with PyTorch API, which allows back-propagation through the module.

\section{Experiment Details}
\subsection{Additional Implementation Details}

\textbf{Example of Training and Evaluation Images.} Figure \ref{fig:widc_example} shows the example of training and evaluation images from one scene. We visualize 16 out of 26 training images.

\noindent \textbf{Matting.} We use the Robust Video Matting approach \cite{li2020robust} with their pre-trained ResNet50 backbone to conduct the matting process. Specifically, we conduct matting on the sequence captured by each camera respectively. Figure \ref{fig:matting} shows an example of matting results. We do observe some artifacts on the matting mask, especially foot regions, which could have some potential impacts on the reconstruction of all approaches.

\noindent \textbf{Scaling.} We normalize scene size to ensure they fit in a unit-sized space. To achieve this, we first estimate size of the human by the reconstruction from the system, and normalize the ray originals and near-far range for ray tracing/sampling.

\noindent \textbf{DTU.} For the DTU dataset, we use the training pipeline and protocol implemented by IDR, which trains 2000 epochs totally on each scene.

\subsection{Implementation Details about Baselines}
We adapt all baselines to performance on the 3D human reconstruction datasets. To be detailed, for those dataset-specific settings and hyper-parameters, we set all approaches to be the same, \eg, pixel selection strategy, ray sampling. We also observe that set the background color to black benefits the volume rendering process on WIDC dataset, thus we use a black background in all experiments on humans. For the method-specific settings and hyper-parameters, \eg, network architecture, positional encoding, number of points sampled per ray, and losses, we use the default setting from each approach. The training batch size depends on the memory of each GPU.

\begin{figure}[t!]
    \centering
    \includegraphics[width=0.47\textwidth]{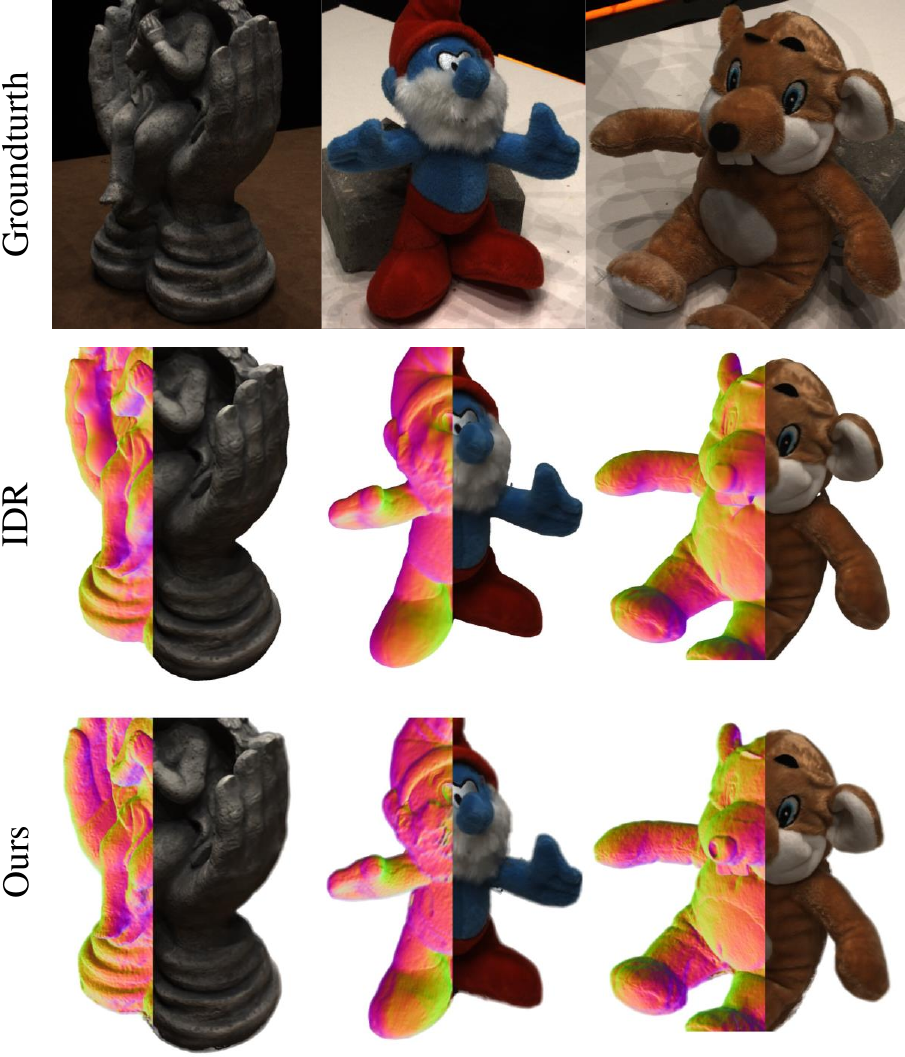}
    \caption{Addtional results on DTU dataset.}
    \label{fig:add_dtu}
\end{figure}

\begin{figure*}[t!]
     \centering
     \begin{subfigure}[b]{\textwidth}
         \centering
         \includegraphics[width=\textwidth]{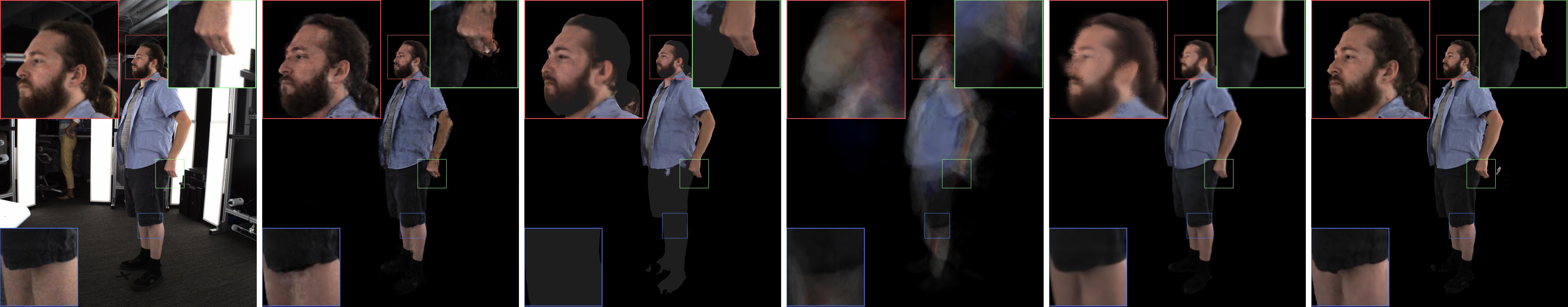}
     \end{subfigure}
     \hfill
     \begin{subfigure}[b]{\textwidth}
         \centering
         \includegraphics[width=\textwidth]{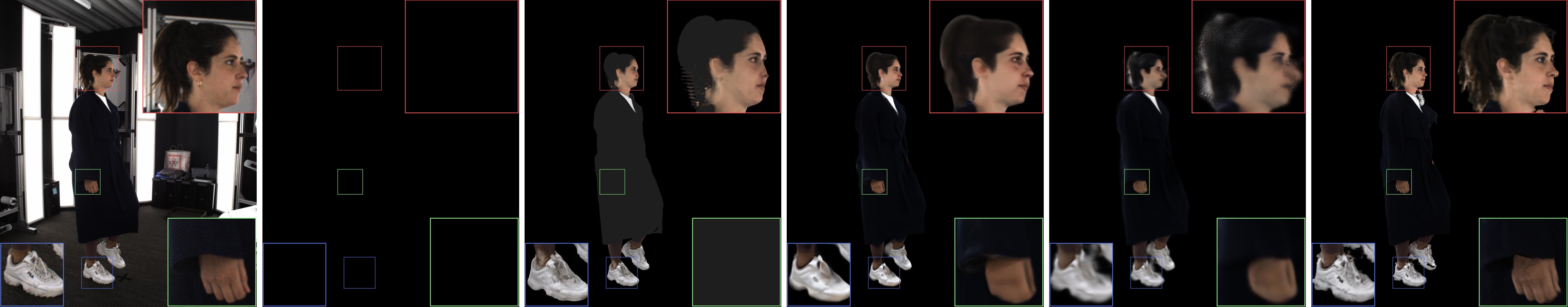}
     \end{subfigure}
     \hfill
     \begin{subfigure}[b]{\textwidth}
         \centering
         \includegraphics[width=\textwidth]{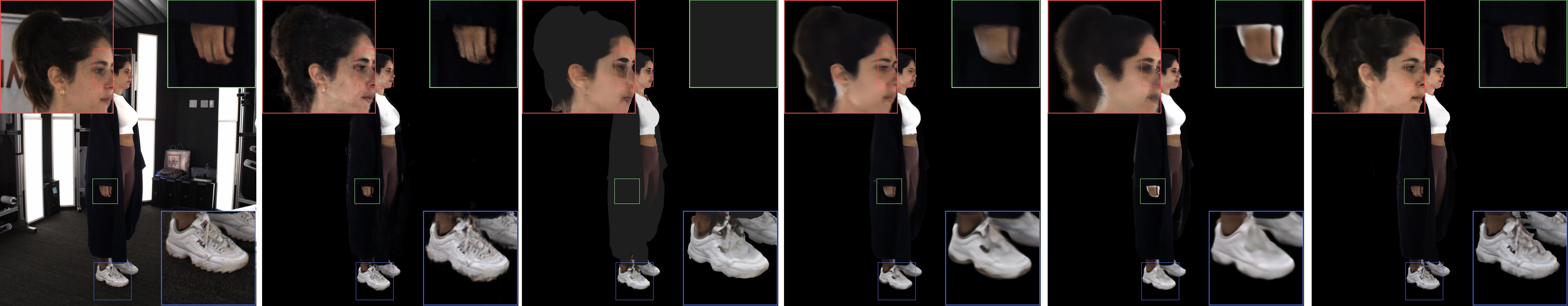}
     \end{subfigure}
     \begin{subfigure}[b]{\textwidth}
         \centering
         \includegraphics[width=\textwidth]{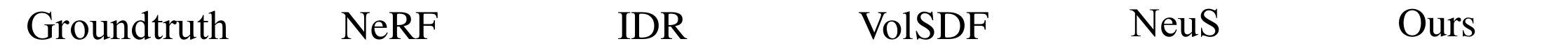}
     \end{subfigure}
    \caption{Additional qualitative results for Novel View Synthesis on WIDC dataset.}
    \label{fig:widc_add_vis}
\end{figure*}

\begin{figure}[t!]
    \centering
    \includegraphics[width=0.47\textwidth]{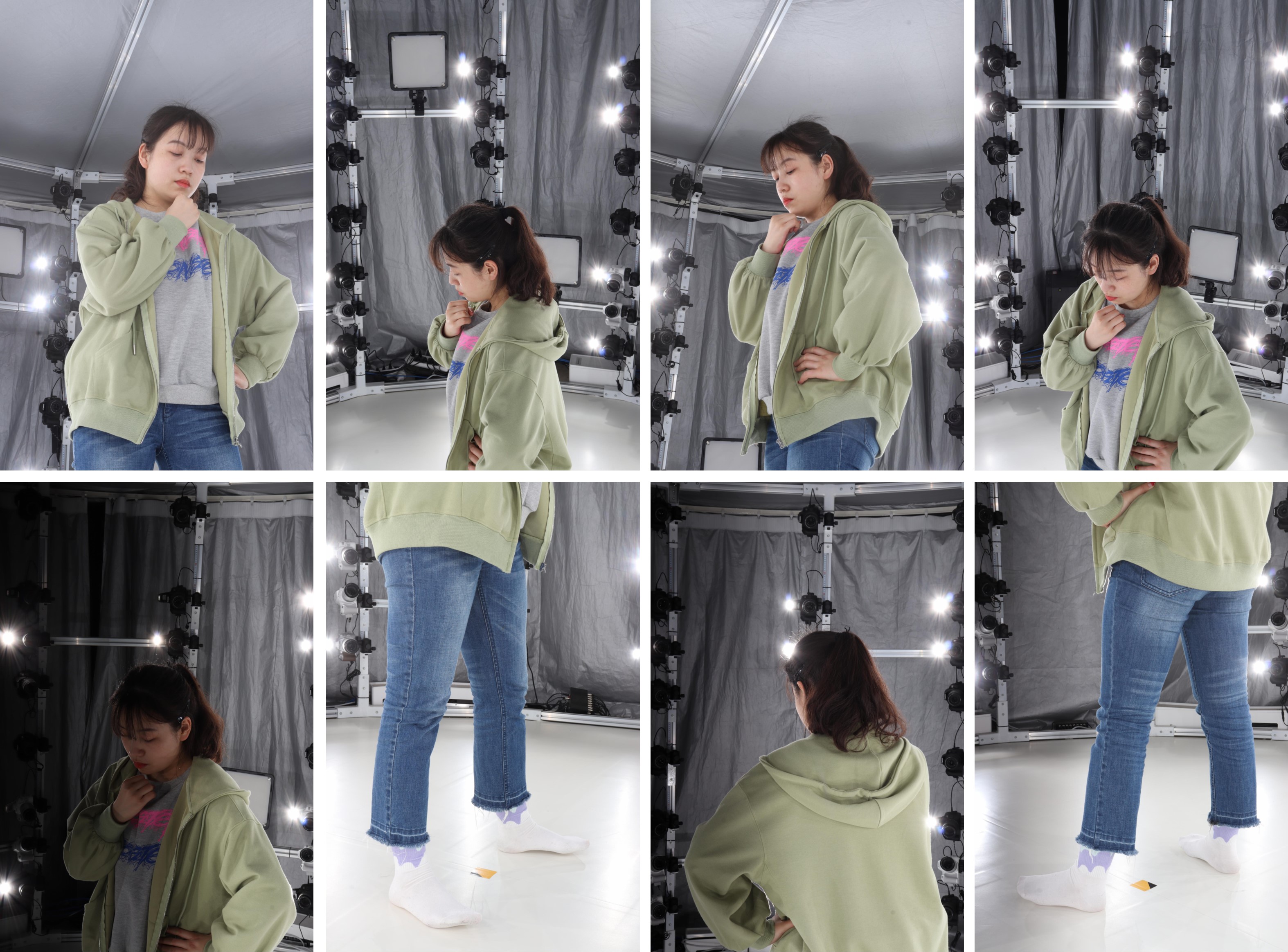}
    \caption{\textit{Example of training image on THUman}}
    \label{fig:train_thuman}
\end{figure}

\noindent \textbf{NeRF.} 
We use the PyTorch implementation of NeRF \cite{lin2020nerfpytorch}. Specifically, the NeRF contains a coarse and a fine rendering network, each network consists of eight layers with 256 channels in each. We use the Adam \cite{kingma2014adam} optimizer with $5e^{-4}$ learning rate and the exponential scheduler to update the learning rate.

\noindent \textbf{IDR.}
We use the official implementation, which includes the mask loss with $weight = 0.01$. The IDR system includes an SDF (8 layers with 512 channels) and an implicit render (4 layers with 512 channels). We find the surface sampling model in the official implementation does not follow the manner described in the paper, but there is no significant difference regarding the reconstruction results.

\noindent \textbf{VolSDF.}
We use the official implementation. However, we find the VolSDF approach is highly likely to fail to converge on some scenes. Thus, we adapt the mask loss used in IDR and NeuS to improve the chance of convergence. Specifically, we set the $mask \ weight = 0.01$. The system includes an 8-layer 256-channel MLP as SDF and a 4-layer 256-channel implicit renderer. Also, we find VolSDF does consume more memory compared to other approaches for each pixel training, thus we set the batch size to 2048 to fit the memory.

\noindent \textbf{NeuS.}
We use the official implementation of NeuS. Specifically, we use the \textit{with mask} setup and set $mask \ weight = 0.1$. The NeuS pipeline consists a density network (8 layers with 256 channels), a SDF network (8 layers with 256 channels), and an implicit renderer (8 layers with 256 channels). We use the default Adam optimizer with $5e^{-4}$ learning rate.

\subsection{Data Capture Protocol}

\textbf{WIDC.} Participants were paid \$100 per hour to be captured in our lab. Participants were instructed to perform a standardized set of movements while varying their levels of dress, which encompassed four conditions: wearing a complete outfit consisting of a jacket, top, and bottom; wearing only their top and bottom without a jacket; wearing only bottoms; and wearing only underwear.

\section{Additional results}

\subsection{Additional Experiments}
\textbf{Error Bar.} Table \ref{tab:error} shows the quantitative comparison for HISR and baselines with error bars.

\noindent\textbf{Visualizations of Reconstructed Meshes.} Figure \ref{fig:normalvsrec} shows the side-by-side comparison of reconstructed surface normals and meshes obtained by marching cubes. All the results shown here are obtained from HISR. The visualization demonstrates, by rendering surface normals, we can have better understanding and evaluation of the reconstructed geometries, thus we chose to visualize surface normals as qualitative comparison for geometry in all experiments.

\noindent\textbf{Additional Results.} Figure \ref{fig:widc_add_vis} shows more qualitative results on the WIDC dataset for novel view synthesis. Figure \ref{fig:add_dtu} shows more qualitative results on the DTU dataset with both surface normal reconstruction and novel view synthesis. Dynamic videos for qualitative results on WIDC are included in the supplementary materials.

\noindent\textbf{HISR on Sequence.} We train HISR on a sequence of the scene. Specifically, we train our model on one out of every eight frames for 4000 epochs and finetune each frame for 1000 epochs respectively. The rendered sequence is in the supplementary materials.
